\documentclass[lettersize,journal]{IEEEtran}
\usepackage{amsmath,amsfonts}
\usepackage{amssymb}
\usepackage{algorithmic}
\usepackage{algorithm}
\usepackage{array}
\usepackage{booktabs}
\usepackage[caption=false,font=normalsize,labelfont=sf,textfont=sf]{subfig}
\usepackage{textcomp}
\usepackage{stfloats}
\usepackage{url}
\usepackage{verbatim}
\usepackage{multirow}
\usepackage{graphicx}
\usepackage{colortbl}
\usepackage{pgfplots}
\usepackage{cite}
\usepackage[normalem]{ulem}
\useunder{\uline}{\ul}{}
\newcommand{\eg}{e.g. }
\newcommand{\etc}{\emph{etc.} }
\newcommand{\ie}{\emph{i.e.} }
\newcommand{\wrt}{\emph{w.r.t.} }

\usepackage[pagebackref,breaklinks,colorlinks]{hyperref}
\usepackage{cleveref}
\definecolor{mygray}{gray}{.9}
\newcommand{\doubleline}[2]{\begin{tabular}[c]{@{}c@{}}{#1}\\ {#2}\end{tabular}}

\begin{document}

\title{CLIPose: Category-Level Object Pose Estimation with Pre-trained Vision-Language Knowledge}

\author{Xiao Lin, Minghao Zhu, Ronghao Dang, Guangliang Zhou, Shaolong Shu,~\IEEEmembership{Senior Member,~IEEE}, Feng Lin,~\IEEEmembership{Fellow,~IEEE}, Chengju Liu and Qijun Chen$^\dagger$,~\IEEEmembership{Senior Member,~IEEE}
\thanks{$^\dagger$ Corresponding author: Qijun Chen (email: qjchen@tongji.edu.cn)}
\thanks{The authors of this paper is supported by the National Natural Science Foundation of China under Grants (62073245, 62233013, 62173248).}
\thanks{X. Lin, M. Zhu, R. Dang, G. Zhou, S. Shu, C. Liu and Q. Chen are with the College of Electronics and Information Engineering, Tongji University, Shanghai 201804, China (email: \{2111118, tjdezmh, dangronghao, tj\_zgl, shushaolong,  qjchen\}@tongji.edu.cn).}
    \thanks{F. Lin is with the College of Electronics and Information Engineering, Tongji University, Shanghai 201804, China, and also with the Department of Electrical and Computer Engineering, Wayne State University, Detroit, MI 48202 USA  (email: flin@wayne.edu).}
\thanks{Color versions of one or more figures in this article are available at https://doi.org/xx.xxxx/TCSVT.20xx.xxxxxxx.}
\thanks{Digital Object Identifier xx.xxxx/TCSVT.20xx.xxxxxxx}

}



\maketitle

\begin{abstract}
    Most of existing category-level object pose estimation methods devote to learning the object category information from point cloud modality.
    However, the scale of 3D datasets is limited due to the high cost of 3D data collection and annotation. Consequently, the category features extracted from these limited point cloud samples may not be comprehensive. This motivates us to investigate whether we can draw on knowledge of other modalities to obtain category information.
    Inspired by this motivation, we propose \textbf{CLIPose}, a novel 6D pose framework that employs the pre-trained vision-language model to develop better learning of object category information, which can fully leverage abundant semantic knowledge in image and text modalities.
    To make the 3D encoder learn category-specific features more efficiently, we align representations of three modalities in feature space via multi-modal contrastive learning.
    In addition to exploiting the pre-trained knowledge of the CLIP's model, we also expect it to be more sensitive with pose parameters. Therefore, we introduce a prompt tuning approach to fine-tune image encoder while we incorporate rotations and translations information in the text descriptions. 
    CLIPose achieves state-of-the-art performance on two mainstream benchmark datasets, REAL275 and CAMERA25, and runs in real-time during inference (40FPS).
\end{abstract}


\section{Introduction}
\label{sec:intro}
\IEEEPARstart{E}fficient and accurate estimation of objects' pose is crucial in numerous practical applications, including augmented reality~\cite{marchand2015pose}, autonomous driving~\cite{chen2017multi} and robotic manipulation~\cite{tremblay2018deep}, \etc
Recent works~\cite{wang2019densefusion,peng2019pvnet,gao2021cloudaae,zhou2021semi,he2021ffb6d,lin2023transpose,cao2023dgecn++} have demonstrated remarkable performance in estimating objects' pose on existing benchmarks~\cite{hinterstoisser2011gradient,brachmann2016uncertainty,xiang2017posecnn}, 
they generally can only handle a few "seen" objects with high-quality CAD models, and sometimes even one instance at a time, which does not adhere to real dynamic and unknown world.
\begin{figure}[htbp]
\includegraphics[width=\columnwidth]{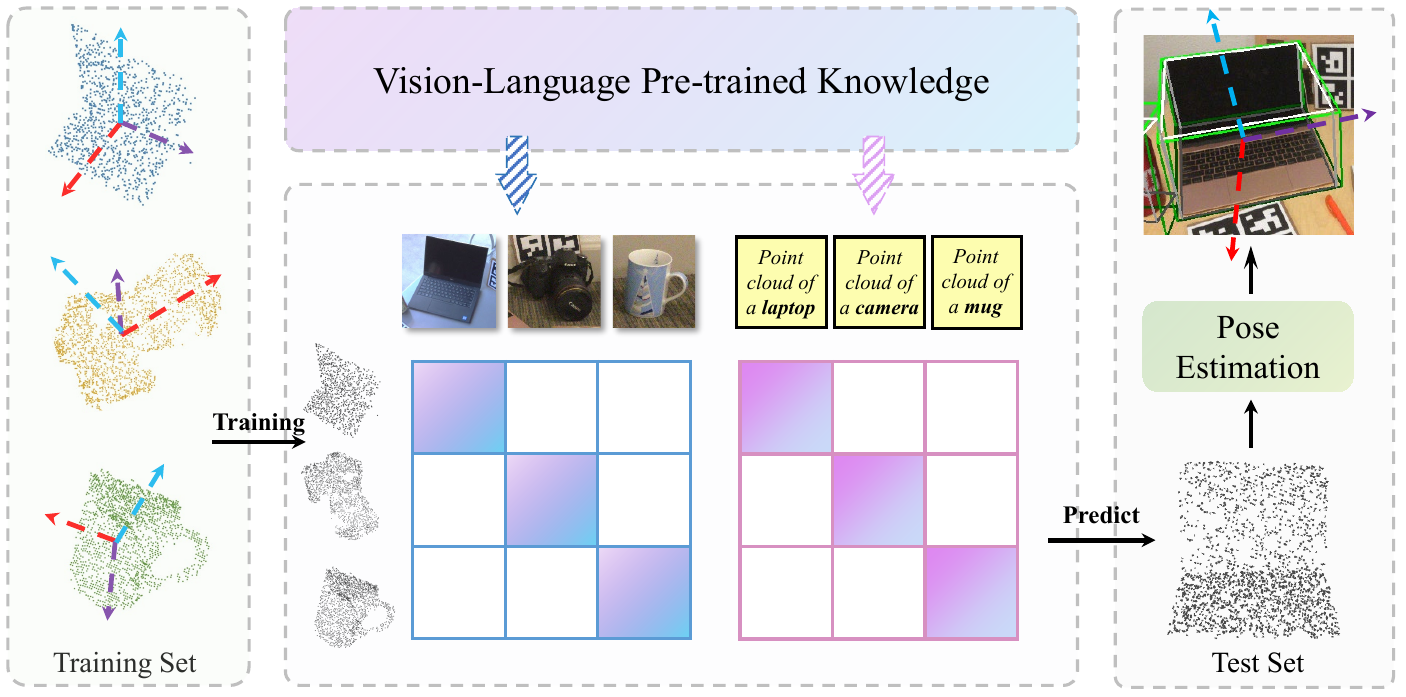}
   \caption{\textbf{Illustration of CLIPose}. The proposed CLIPose takes as input a triplet of objects (point cloud, image and text), and aligns the representations of three modalities in feature space via contrastive learning. This enables the network to obtain more robust category-specific information from vision-language pre-trained knowledge. }
   \label{fig:head_fig}
\end{figure}
In recent years, category-level object pose estimation methods~\cite{wang2019normalized,tian2020shape,chen2021fs,di2022gpv,liu2023prior,zou2023gpt} have emerged, whose objective is to estimate the pose of previous "unseen" objects within predefined categories. 
However, developing such algorithms is inherently more challenging due to the complicated and diverse intra-category variation, including shape and texture variation.

To address the above-mentioned issue, a straightforward idea is to extract category-specific feature from 3D point cloud data, as the objects of the same category typically exhibit similar geometric characteristics. 
For instance, mugs are typically cylindrical with a handle while laptops generally come with hinged structures.
Most of previous category-level methods conducted research following the idea mentioned above.
\cite{tian2020shape} extracts the point cloud mean shape of each category, referred to as \emph{shape prior}, while~\cite{di2022gpv} indirectly acquires category information by predicting the objects' bounding boxes. 
Although the positive outcomes of existing methods, there is still room for improvement in two aspects: 
1) These methods heavily depend on complicated manual designs or various intermediate processes to obtain category information.
2) The category information extracted from point cloud is limited by the scale of existing category-level pose estimation datasets due to the prohibitive cost of 3D data collection and annotation~\cite{goyal2021revisiting,wu20153d,yu2022point}.
Therefore, it's essential to explore more effective methods to extract richer category-specific feature for better pose estimation performance.

More recently, multi-modal learning~\cite{radford2021learning,li2021align,yu2022coca,dang2023instructdet} has advanced rapidly and drawn great attention. 
Numerous proposed models, represented by CLIP~\cite{radford2021learning} exploits language knowledge to assist vision tasks and demonstrate promising outcomes~\cite{qiu2021vt,li2022grounded,li2022language}. 
These works indicate that leveraging knowledge from different modalities can effectively promote semantic understanding in the original modality. 
It motivates us to employ rich knowledge from image and text modalities in category-level pose estimation task. 
Some recent methods~\cite{zhang2022pointclip,xue2023ulip} have explored how multi-modal information can assist 3D data understanding. 
However, these works primarily focus on point cloud classification tasks. As for whether multi-modal learning can further facilitate other downstream 3D tasks, \eg object pose estimation, is still not well studied.

In this paper, we propose a concise yet effective category-level pose estimation model, named CLIPose, which leverages pre-trained vision-language knowledge to assist in extracting more robust category-specific feature for pose estimation. 
The illustration of CLIPose is shown in \cref{fig:head_fig}.
Specifically, we first acquire three modalities features of the target objects: 3D point clouds, image patches and text descriptions that specify its category.
Subsequently, the features obtained from three modalities are aligned through the contrastive learning.
The alignment operation can reduce the distance between different modalities in the feature space, which enables more comprehensive point cloud feature learning.
Unlike directly learning with other data modalities~\cite{fan2021acr,lin2022category} (\eg RGB), making use of a pre-trained vision-language model allows us to utilize abundant semantics extracted from image-text feature space to capture richer category-specific information for category-level pose estimation task.
Furthermore, we expect CLIP to be aware of the pose information of objects in addition to the category feature. 
Hence, we introduce a prompt tuning approach that can update the parameters of image encoder based on pose estimation results during pre-training. 
Moreover, we innovatively create text descriptions with pose parameters, which can further assist the network to obtain pose-sensitive information.

Our primary contributions are summarized as follows:
\begin{itemize}
    \item We propose a novel category-level 6D object pose estimation framework, CLIPose. To the best of our knowledge, our method is the first work that employs the pre-trained vision-language model of CLIP~\cite{radford2021learning} to the category-level pose estimation task, which can leverage abundant semantic knowledge in image and text modalities to extract more robust category-specific feature.
    \item We align the representations from three modalities in feature space via contrastive learning. This allows the network to learn object's category information in a concise manner, eliminating the need for shape prior or complicated manual design. Meanwhile, we finely design a prompt tuning approach to fine-tune the image encoder, which enables the CLIP model to better adapt to pose estimation task.
    \item We provide a brand new insight to realize the real open-world object pose estimation. Our method achieves state-of-the-art or competitive results on two mainstream benchmarks, REAL275 and CAMERA25, which demonstrates the effectiveness of CLIPose.

\end{itemize}

The rest of the paper is organized as follows. Section \ref{sec:relate_works} reviews several previous works on category-level object pose estimation and Multi-modal Representation Learning. 

The alignment among different modalities and the prompt fine-tuning approach for CLIP image encoder are detailed in Section \ref{sec:Methodology}. Finally, we report and analyze the experimental results in Section \ref{sec:Experiments}, and conclude our paper in Section \ref{sec:Conclusion}.

\section{Related Works}
\label{sec:relate_works}
\subsection{Category-level Object Pose Estimation}
Category-level object pose estimation methods~\cite{wang2019normalized,tian2020shape,chen2021sgpa,fan2021acr,di2022gpv,zheng2023hs,liu2023prior} consist of two major lines of research. One line concentrates on the correspondence between point clouds. 
NOCS~\cite{wang2019normalized} suggests mapping input shape to a normalized canonical space and recovering the pose via Umeyama algorithm~\cite{umeyama1991least}. 
SPD~\cite{tian2020shape} trains a shape deformation field and leverages the structural similarity between shape prior and target object to estimate the pose. 
ACR-Pose\cite{fan2021acr} proposes an generative adversarial training network to acquire more accurate NOCS models.
while SGPA\cite{chen2021sgpa} utilizes Transformer\cite{vaswani2017attention} for global similarity modeling of shape priors and target objects. 
Though effective, these methods suffer from the high compute complexity of point cloud matching.
The other line of research explores how to achieve end-to-end pose estimation without explicit point cloud alignment. 
CASS\cite{chen2020learning} estimates the object pose via the learning of a canonical shape space implicitly.
GPV-Pose~\cite{di2022gpv} utilizes point-wise bounding box voting to capture category-specific feature, while HS-Pose~\cite{zheng2023hs} proposes hybrid scope feature extractor to substitute the 3D-GCN module in GPV-Pose.
IST-Net~\cite{liu2023prior} demonstrates the shape prior is not the credit to achieve precise pose estimation but shape deformation field, then proposes an implicit space transformation to recover pose parameters.
The above-mentioned works concentrate on acquiring category information from depth or RGB-D data. However, the category features extracted from limited 3D data may not be well-suited to address the complicated and diverse intra-category variations.
On the contrary, we propose a novel framework that exploits pre-train knowledge from image and text modalities to extract more comprehensive category-specific features.


\begin{figure*}[htbp]
    \includegraphics[width=\textwidth]{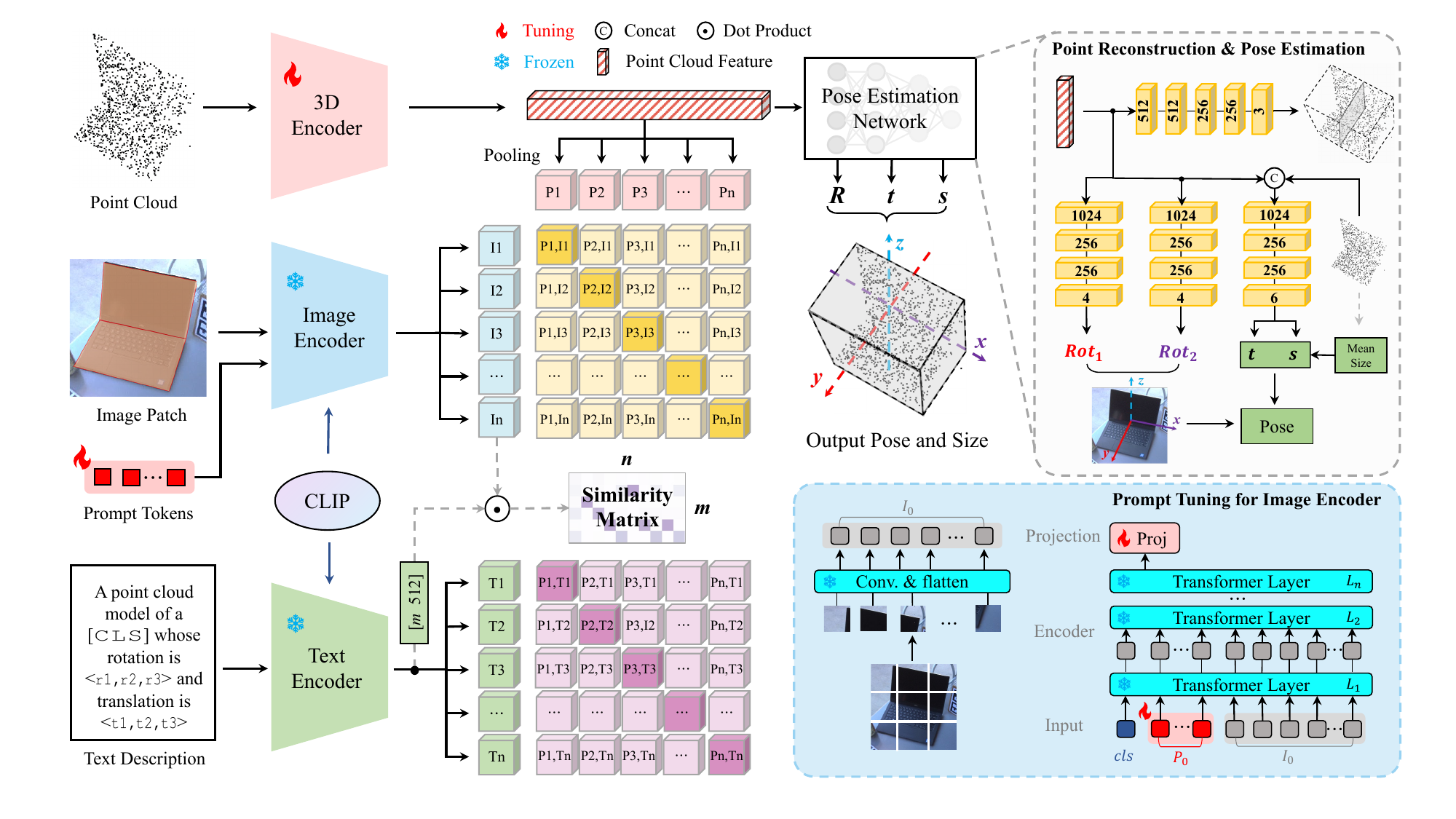}
    \caption{\textbf{Overview of CLIPose.} The inputs of our framework are a batch of objects (\eg \emph{laptop}) represented as triplets (point cloud, image, text), therein we append rotation and translation information to the text description. Image and text features are extracted from a pre-trained (\textcolor{blue}{Frozen}) vision and language model, and point cloud features are obtained by a 3D encoder (\textcolor{red}{Tuning}). Contrastive losses are applied to align the 3D representations of an object to its image and text representations during pre-training. We utilize CLIP's inherent classification ability to predict the category of the input object (\emph{m} is the number of categories) to form the classification loss.
    The image encoder would be fine-tuned with additional prompt tokens \textcolor{red}{$P_0$} (\textbf{Bottom Right}). 
    }\label{fig:CLIPose}
\end{figure*}

\subsection{Multi-modal Representation Learning}
Most of existing multimodal learning~\cite{radford2021learning,li2021align,yu2022coca,dang2023instructdet} methods are about image and text modalities. 
Among these works, CLIP~\cite{radford2021learning} is the  pioneering effort that utilizes image and text encoders to extract feature respectively for each image-text pair, and then aligns the features from both modalities via contrastive learning. 
Thanks to the outstanding success of CLIP, various image-text related works draw the idea from CLIP, applying contrastive learning architecture or modal alignment approach in semantic segmentation~\cite{li2022language,xu2022groupvit}, 
vision-language navigation~\cite{wang2023res,he2023learning} 
and video retrieval~\cite{zhu2023fine}, \etc. 
For the 3D vision domain, some recent works~\cite{zhang2022pointclip,zhu2022pointclip,zhang2022can,xue2023ulip,xue2023ulip2} explore the application of multimodal representation learning and present promising results.
The PointCLIP series~\cite{zhang2022pointclip,zhu2022pointclip} converts 3D piont cloud to multi-view 2D depth maps and employs CLIP directly for 3D recognition.
DepthCLIP~\cite{zhang2022can} cleverly transforms the depth estimation problem into a the classification problem via classifying the depth of objects into seven classes from near to far.
The ULIP series~\cite{xue2023ulip,xue2023ulip2} learns the unified representation with the goal of developing a 3D version of CLIP.

However, above methods typically focused on point cloud classification tasks. The application of multimodal learning and CLIP's robust pre-trained knowledge in other 3D downstream tasks, \eg object pose estimation, is still not further explored.
To this end, our method develops the contrastive learning framework among three modalities, utilizing the pre-trained knowledge from image and text modalities to extract more robust category-specific feature for accurate category-level object pose estimation.

\section{Methodology}
\label{sec:Methodology}

\subsection{Preparing Triplets}
\label{sec:triplets}
In datasets of category-level pose estimation, the image patch and point cloud of a specific object are typically corresponding. The text descriptions can be obtained according to the category of objects.
Naturally, we can treat the representations from these three modalities as a triplet.

For each object $i$ in dataset, we prepare the triplet containing image patch, point cloud and text description. 
Subsequently, we leverage these triplets to train our network. 
During each iteration of training, we randomly select a triplet and take it as input of three modalities encoder to extract features,
\begin{equation}
\label{equ:encoder}
    \mathbf{f}_{i}^{M}= \Phi_{M}(M_{i})
\end{equation}
where $M$ indicates modalities, $\mathbf{f}_{i}^{M}$ denotes the features extracted from specific modalities and $\Phi_{M}$ represents feature encoder of modalities.

{\bf Image Patch of Object.} 
We first employ the off-the-shelf Mask-RCNN~\cite{he2017mask} object detector to segment the object region of interests(ROIs) from RGB images. 
In general, the ROIs only contain the target object, so it can also be referred to as an object image patch.
Then, all object images patches after normalization are fed into the image encoder of CLIP, which is a standard vision transformer(ViT)~\cite{dosovitskiy2020image} in this paper.
The output features of image encoder will be involved in the formation of two loss functions. One is the alignment with the point cloud features to form the contrastive learning loss, and the other is the similarity computation with the text features to get the loss for predicting the category of objects.

{\bf Point Cloud Generation. }
Given a RGBD image, we can obtain the 2D pixel coordinates $(u,v)$ of the target object.
According to the pinhole camera model, given the determined camera intrinsics $C$, the correspondence between the 3D point coordinates $[X,Y,Z]$ and the 2D pixel coordinates $[u,v]$ in the image is as follows,
\begin{equation}
\label{equ:2d_2_3d_matrix}
    \begin{bmatrix}
    X\\Y\\Z
    \end{bmatrix}
    =     C^{-1}Z
    \begin{bmatrix}
    u\\v\\1
    \end{bmatrix},C = \begin{bmatrix}
                    f_{x} & 0 & c_{x}\\0 & f_{y} & c_{y}\\0 & 0 & 1
                    \end{bmatrix}
\end{equation}
where the $Z$ coordinate indicates the depth of the corresponding 2D pixel point.
We can leverage the RGBD images to construct the point cloud of each target object based on \cref{equ:2d_2_3d_matrix}.
The point cloud features extracted by 3D encoder will be aligned with the image features and text features respectively via contrastive learning.
In addition, we utilize the point cloud features for point cloud symmetry reconstruction and final pose estimation.

{\bf Text Descriptions.}
In category-level pose estimation task, the category range of the target object is already predefined. Hence, we could utilize predefined category that come with each object as the corresponding text description.
Prior works~\cite{zhang2022pointclip,xue2023ulip} adopt simple prompts to construct meaningful sentences, such as "a point cloud model of [CLS]", with the aim of enhancing the adaptability of the CLIP model to 3D data. 
However, these sentences only contain objects' category information, which does not guide the network to learn pose-sensitive information.

Therefore, to enable the CLIP model more adapted to the pose estimation task, we leverage the objects' pose parameters to construct text descriptions.
We extract the rotation (euler angles) and translation parameters directly from ground truth.
Thus the format of text description for pre-training can be expressed as "a point cloud model of a [CLS] whose rotation euler angles is $<$-103.73, 58.98, 71.62$>$ and translation is $<$10.63, 48.78, 41.32$>$."
The process of generating text descriptions would be done offline and does not impact the training efficiency of the network.
During the inference stage, the object's pose parameters are required to be recovered, so the text format is as follows: "a point cloud model of a [CLS] whose pose is to be estimated."

\subsection{Aligning Representations}
\label{sec:aligning}
With the obtained features of three modalities: point cloud, image and text, CLIPose performs pre-training to align all modal representations into the same feature space via contrastive learning.
We draw the inspiration from the training strategy of CLIP and achieve alignment by calculating the similarity between different modalities.
Specifically, we calculate the cosine similarity between point cloud and image modalities, as well as between point cloud and text modalities.
Then, the contrastive learning loss is computed by the cross entropy between the obtained similarity matrix and the ground truth. The ground truth is an identity matrix, which represents stronger corresponding for the same object.
Moreover, since the scale of 3D point cloud data is limited, we find that catastrophic forgetting will emerge if we full-tune and update the pre-aligned image-text representations of CLIP during training.
Hence, we freeze majority of CLIP's model parameters and train the 3D point cloud encoder following the aforementioned contrastive learning approach~\cite{xue2023ulip,zhang2022can}. 
A small portion of image and text encoder weights are reserved for fine-tuning, which is detailed in \cref{sec:prompt_tune}.

The alignment operation between different modalities serves two crucial functions.
First, aligning representations of all modalities into the same feature space enhances the consistency among different modalities.
Second, the wealth of semantics information already captured by CLIP can effectively assist the feature learning of 3D encoder, which enables better 3D understanding, thereby improving the performance in category-level pose estimation tasks.

{ \bf Multi-Modal Contrastive Learning Loss.} 
The contrastive learning is initially developed to address multi-categorical Contrastive Estimation (CE) problems.
The ideal contrastive loss is a function whose value will be low when query is similar to its positive key (correct category) and dissimilar to negative keys (wrong categories), where the cross entropy function is a commonly contrastive learning loss function. 
\begin{figure}[t]
    \includegraphics[width=\columnwidth]{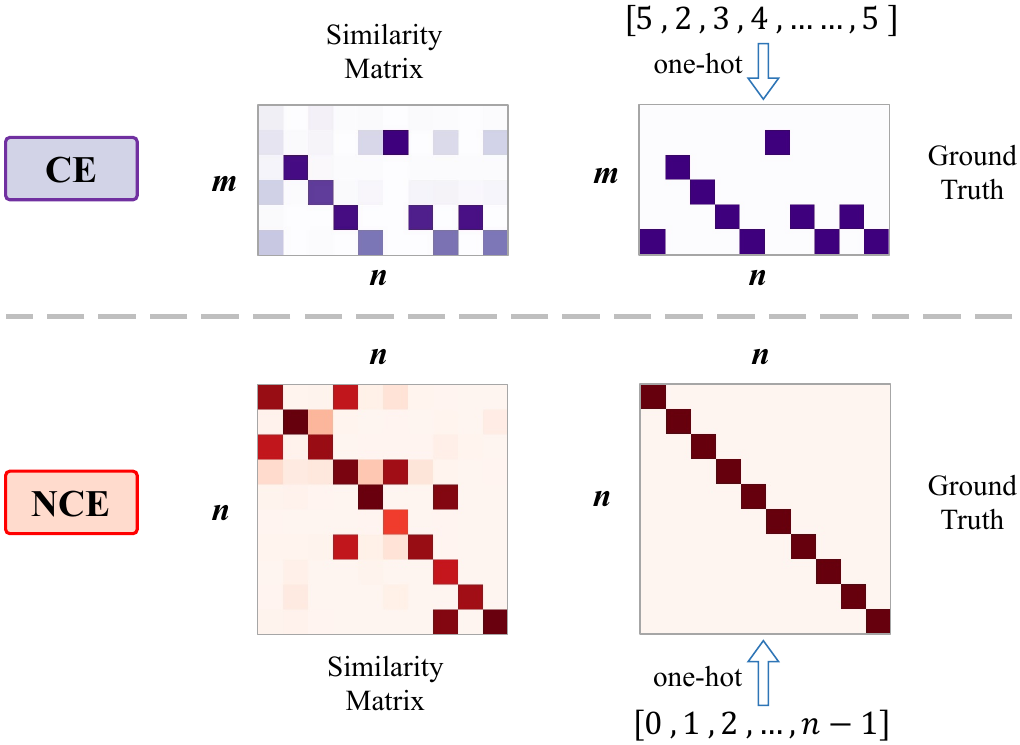}
    \caption{\textbf{Comparison of the loss function in CE and NCE.} The darker the color means the higher similarity value. $m$ denotes the number of predefined categories. $n$ indicates the number of input samples. The ground truth list [···] for each input batch could be formed into a similarity ground truth matrix using one-hot encoding.}
   \label{fig:cevsnce}
\end{figure}
Typically, the cross entropy loss is suitable for multi-categorical problems and it strives to classify each sample accurately.
Apart from accurate classification results, our method is more concerned with increasing the consistency of the three modal representations via contrastive learning. Directly using the conventional cross entropy loss does not meet our requirements effectively.
Draw the idea of prior work~\cite{he2020momentum}, we treat the alignment among different modalities as a Noise Contrastive Estimation (NCE)~\cite{dyer2014notes} problem and design the multi-modal noise contrastive loss for our framework.
The comparison between the loss function used in CE and NCE is shown in \cref{fig:cevsnce}.

The essence of NCE lies in transforming a multi-categorical problem into a binary-categorical problem. 
If the representations of the two modalities are matched, this sample is considered positive sample and the rest are negative samples (noise samples).
Hence, the ground truth of noise contrastive loss is a unitary matrix, as shown in bottom right of \cref{fig:cevsnce}. 
Through learning the difference between positive and negative samples, the network is able to learn the correspondence between two modalities.
With the similarity matrices measured by dot product, we compute the contrastive loss for each pair of modalities as follows,
\begin{equation}
\label{equ:multimodal_loss}
\begin{split}
    \mathcal{L}_{(M_{1},M_{2})}^{NCE}= \sum_{(i,j)}-\frac{1}{2}log\frac{exp( \mathbf{f}_{i}^{M_1}\cdot\mathbf{f}_{j}^{M_2}/\tau )}{\sum_{k=1}^{K}exp(\mathbf{f}_{i}^{M_1}\cdot\mathbf{f}_{k}^{M_2}/\tau )}\\-\frac{1}{2}log\frac{exp(\mathbf{f}_{i}^{M_1}\cdot\mathbf{f}_{j}^{M_2}/\tau )}{\sum_{k=1}^{K}exp(\mathbf{f}_{k}^{M_1}\cdot\mathbf{f}_{j}^{M_2}/\tau )}
\end{split}
\end{equation}
where $M_{1}$ and $M_{2}$ indicate two different modalities, $\mathbf{f}_{i}^{M_1}$ denotes the $i_{th}$ feature extracted from $M_{1}$ modality, $\tau$ is a temperature hyper-parameter, $(i,j)$ represents a positive pair in each training batch, $K$ is the number of negative samples and $k$ indicates each negative sample from 1 to $K$. 
Eventually, we minimize all modality pairs of contrastive loss as follows, 
\begin{equation}
\label{equ:all_contrastive_loss}
    \mathcal{L}_{mm}^{NCE}= \alpha \mathcal{L}_{(PC,img)}^{NCE}+\beta \mathcal{L}_{(PC,text)}^{NCE}+\lambda 
 \mathcal{L}_{(img,text)}^{NCE}
\end{equation}
where $PC$ indicates point cloud modality. By default, $\alpha$ and $\beta$ are set to be constant 1.0 equally, and $\lambda$ is set to be 0; 
because the image and text modalities are pre-aligned in CLIP, these two modalities already possess a high level of consistency. 
Since we freeze most of the parameters of CLIP's encoder, the similarity values of the NCE loss between these two modalities consistently remain at low levels and contributes very little to the updates of the network parameters.

We initially intend to leverage the the first two terms of the \cref{equ:all_contrastive_loss} directly.
However, the association between image and text contains the most intuitive information about object categories, which deserves to be explored further.
In zero-shot classification tasks, CLIP computes the text and image representations by dot product to obtain the similarity and predicts the category of objects.
This suggests that we can leverage CLIP's inherent categorization capabilities to further enhance the network's learning of category features.
Similar to the inference stage of CLIP, we calculate the similarity matrix by dot product of text and image representations to obtain preliminary classification results for each object.
Subsequently, we construct the multi-categorical Contrastive Estimation (CE) loss by computing the cross entropy function as follows, 
\begin{equation}
\label{equ:category_loss}
    \mathcal{L}_{(img,text)}^{CE} = -\sum_{(i,j)}log\frac{exp( \mathbf{f}_{i}^{img}\cdot\mathbf{f}_{j}^{text})}{\sum_{m=1}^{M}exp(\mathbf{f}_{i}^{img}\cdot\mathbf{f}_{m}^{text} )}
\end{equation}
where $CE$ indicates that we utilize the normal cross entropy loss, and $M$ donates the number of categories. 

In this way, we enhance the consistency in representations between the point cloud modality and the image-text modality, while simultaneously introducing classification results through contrastive estimation.
We conduct ablation experiments \cref{sec:ablation_study} to validate the effectiveness of proposed loss.


\subsection{Prompt Fine-tuning for CLIP}
\label{sec:prompt_tune}
The original CLIP model is primarily designed to solve image classification problems. Consequently, CLIP is highly sensitive to category-related features of objects, while the perceptive of object pose related information is comparatively less sharp.
While fully leveraging the rich semantic knowledge embedded in the CLIP encoders, we expect that CLIP can be more attuned to the 6D pose of objects.
An intuitive idea is to fully update the parameters of CLIP image and text encoder along with the 3D encoder. Nevertheless, as mentioned earlier, this way would lead to catastrophic forgetting.
If the parameters are completely frozen, the CLIP model is unable to effectively obtain pose information.
Hence, exploring a appropriate and efficient fine-tune approach is crucial to serve both purposes. 

We first start with the image encoder.
The CLIP's image encoder is a standard ViT~\cite{dosovitskiy2020image}, which initially reshape the image $I \in \mathbb{R}^{H \times W \times C}$ into a sequence of flattened 2D image patches $I_{0} \in \mathbb{R}^{N \times (P^{2} \cdot C)}$, where $(H,W)$ denotes the resolution of the original image, $C$ is the number of channels, $(P,P)$ is the resolution of each image patch, and $N=HW/P^2$ is the resulting number of patches. 
Then, ViT prepends a learnable embedding [$cls$] token to the sequence of embedded patches ($I_{0}^{0}=[cls]$), serving as the final input.
The [$cls$] token captures the category information of the image through multiple Transformer layers, which is then mapped into specific dimensions by a projection layer and utilized for representations alignment.
Recalling the two main purposes of this paper. We aim to utilize the abundant pre-training knowledge to assist robust category-specific feature learning and expect the CLIP model to be more pose-sensitive. 
The output image features are mainly related to the input image embedding. Naturally, we can explore fine-tuning in this aspect.
Drawing the idea from \cite{jia2022visual}, we present a prompt fine-tuning manner to appropriately update the parameters of the CLIP's image encoder during training, as shown in the bottom right of the \cref{fig:CLIPose}. 
Specifically, we prepend a set of prompt tokens with the same dimensions as the input image embedding, then we insert prompt tokens $P_{0}\in \mathbb{R}^{N \times (P^{2} \cdot C)}$ between the [$cls$] embedding and sequence of embedded image patches $I_{0}$ as follows,
\begin{equation}
\label{equ:prompt_prepend}
    \mathcal{I}_0 = [cls;P_0;I_{0}^{1};I_{0}^{2};\cdots;I_{N}^{1}]
\end{equation}
where $\mathcal{I}_0$ is the integrated embedding that would be fed into the image encoder. The [$cls$] embedding outputs the final image features through the projection layer.
Since the default length of an image embedding is 196, we select 10 as the length of the prompt tokens to strike the balance. This length won't affect the original image features while facilitating effective fine-tuning.
To retain pre-trained knowledge of the image modality, we freeze all parameters of image encoder except for the prompt tokens $P_{0}$ and the parameters of projection layer.

As for the text encoder, the incorporation of pose parameters and text description (\cref{sec:triplets}) has already provided pose-sensitive information for text modality, so we do not further fine tune the text encoder.
We also take into account various factors that might affect prompt fine-tuning, including the length of prompt tokens, insertion location and the number of involved transformer layers, \etc. 
Extensive ablation experiments \cref{sec:ablation_study} are performed to validate the effect of these factors on fine-tuning.


\subsection{Pose Estimation and Overall Loss Function}
\label{sec:estimation_and_loss}
Given the triplets of target objects, the objective of CLIPose is to estimate the pose of input object, including rotation $R \in SO(3)$, translation $t \in \mathbb{R}^3$ and $s \in \mathbb{R}^3$. 
We utilize point cloud features extracted from 3D encoder to recover object 6D pose parameters, as shown in the top right of the \cref{fig:CLIPose}. 
We adopt the pose estimator in GPV-Pose~\cite{di2022gpv}, which decompose the rotation into three mutually perpendicular bounding box plane normals $\left [r_{x},r_{y},r_{z}  \right ]$.
We predict the first two normals ($r_{x}$ and $r_{y}$), and the last rotation $r_{z}$ is available by cross product. 
Note that we calibrate the plane normals ($r_x$, $r_y$) to be perpendicular normals ($r_{x}^*$, $r_{y}^*$) before calculating the loss by minimize the following cost function,
\begin{equation}
\label{equ:sullp_to_perpend_normals_1}
\begin{split}
    \theta_{1}^{*},\theta_{2}^{*}&=arg \ min \ c_{y}\theta_{1}+c_{x}\theta_{2}\\
    s.t.\ \theta&=\theta_{1}+\theta_{2}+\pi/2,
\end{split}
\end{equation}
where $\theta$ refers to the angle between origin plane normals and $c_x$ and $c_y$ denote the confidence of $r_x$ and $r_y$. From \cref{equ:sullp_to_perpend_normals_1} we can then obtain
\begin{equation}
\label{equ:sullp_to_perpend_normals}
\begin{split}
      \left\{\begin{matrix}
    \theta_{1}^{*}=\frac{c_x}{c_x+c_y}( \theta-\frac{\pi}{2})\\\theta_{2}^{*}=\frac{c_y}{c_x+c_y}( \theta-\frac{\pi}{2}).
    \end{matrix}\right.
\end{split}
\end{equation}
The calibrated plane normals ($r_{x}^*$, $r_{y}^*$) can be calculated from ($\theta_{1}^{*},\theta_{2}^{*}$) via the Rodrigues Rotation Formula~\cite{rodrigues1840lois}. 
Therefore, the rotation can be obtained as $\left [r_{x}^{*},r_{y}^{*},r_{z}^{*} \right ]$, and the rotation loss as follow,
\begin{equation}
\label{equ:sullp_rotation_loss}
    \mathcal{L}_{rot} = \left\|r_x - r_{x}^{gt}\right\|_{1}+\left\|r_y - r_{y}^{gt}\right\|_{1}
\end{equation}
where $r_{x}^{gt}$ and $r_{y}^{gt}$ indicate the ground truth plane normals, $\left\|\cdot \right\|_1$ refers to the $\mathcal{L}_{1}$-loss.

As for translation $t$, which can be computed with the residual prediction $t_{*}$ and the mean of input point cloud $M_{P}$ as $t=t_{*}+M_{P}$. 
Thus we can calculate the translation loss as,
\begin{equation}
\label{equ:sullp_translation_loss}
    \mathcal{L}_{trans} = \left\|t-t^{gt} \right\|_1
\end{equation}
where $t^{gt}$ and $s^{gt}$ donates the ground truth translation.
Similarly, the scale $s$ can be obtained as $s=s_{*}+M_{s}$, where $s_*$ is the residual scale and $M_s$ indicates the mean object scale of all instances within a certain category in the training dataset.
The scale loss is defined as,
\begin{equation}
\label{equ:sullp_scale_loss}
    \mathcal{L}_{scale} = \left\|s-s^{gt} \right\|_1
\end{equation}
where $s^{gt}$ donates the ground truth scale.
Finally, the total pose loss term can be expressed as follows,
\begin{equation}
\label{equ:pose_loss}
\begin{split}
     \mathcal{L}_{pose} = \mathcal{L}_{rot} + \mathcal{L}_{trans} + \mathcal{L}_{scale} \\
\end{split}
\end{equation}

\begin{table*}[htbp]
    \small
    \centering
    \setlength\tabcolsep{11pt}
    \renewcommand\arraystretch{1.2} 
    \caption{\textbf{Comparisons with state-of-the-art methods on REAL275 dataset.}  Overall best results are in bold. \textbf{Input} denotes the format of the input data used by the methods, and \textbf{Prior} refers to whether the method necessitates prior category information.
    ‘-’ indicates no results reported under this metric in original paper.
    ($\uparrow$) represents a higher value indicating better performance.
    }
    \begin{tabular}{r|cc|ccccc|c}
    \toprule[1.2pt]
     Methods & Input & Prior  & 5°2\emph{cm}$\uparrow$ & 5°5\emph{cm}$\uparrow$ & 10°2\emph{cm}$\uparrow$ & 10°5\emph{cm}$\uparrow$ & 10°10\emph{cm}$\uparrow$ & Speed(FPS)$\uparrow$\\
    \midrule
    NOCS\cite{wang2019normalized}     &  RGB-D  & $\times$        & -         & 9.5        & 13.8       & 26.7             & 26.7      & 9.9          \\
    CASS\cite{chen2020learning}     &  RGB-D  & $\times$        & 19.5         & 23.5        & 50.8       & 58.0             & 58.3        & -          \\
    DualPoseNet\cite{lin2021dualposenet}   &  RGB-D  & $\times$ & 29.3         & 35.9        & 50.0       & 66.8             & -        & 2.6          \\
    SPD\cite{tian2020shape}     &  RGB-D  & \checkmark        & 19.3         & 21.4        & 43.2       & 54.1             & -        & 7.9          \\
    SGPA\cite{chen2021sgpa}   &  RGB-D  & \checkmark        & 35.9 & 39.6         & 61.3 & 70.7           & - & 6.6                                  \\
    DPDN\cite{lin2022category}    &  RGB-D  & \checkmark & 46.0 & 50.7         & \textbf{70.4} & 78.4           & 80.4 & 31.6   \\
    IST-Net\cite{liu2023prior}    &  RGB-D  & $\times$    & 47.8         &55.1        & 69.5        & 79.6      & 81.6        & 33.7          \\
    \hline
    FS-Net\cite{chen2021fs}   &  D  & $\times$ & 19.9         & 33.9        & -      & 69.1             & 71.0        & 16.1          \\
    GPV-Pose\cite{di2022gpv}    &   D  & $\times$ & 32.0 & 42.9         & 55.0 & 73.3           & 74.6 & 29.6                                   \\
    SSP-Pose\cite{zhang2022ssp}    &   D  & \checkmark & 34.7 & 44.6         & - & 77.8           & 79.7 & 37.0                                   \\
    GPT-COPE\cite{zou2023gpt}    &   D  & \checkmark & 45.9 & 53.8        & 63.1 & 77.7           & 79.8 & -                                   \\
    DR-Pose\cite{zhou2023dr}    &  D  & \checkmark         & 41.7 & 46.0         & 67.7 & 76.3           & - & -                                   \\
    HS-Pose\cite{zheng2023hs}    &   D  & $\times$  & 45.3          &54.9         & 68.6         & 83.6      & 84.6         & {\ul39.8}          \\
    \midrule
    Ours    &  D  & $\times$ & \textbf{48.7}         & \textbf{58.3} & \textbf{70.4} & \textbf{85.2} & \textbf{86.2} & \textbf{40.0} \\
    \bottomrule[1.2pt]
    \end{tabular}
    \label{tab:compare_sota}
\end{table*}

The pose estimation network actually includes a point cloud symmetry reconstruction component, which is designed to extract more accurate point clouds for reflection (\emph{mug,laptop}) and rotational (\emph{can,bowl,bottle}) symmetry objects.
We also supervise it by $\mathcal{L}_{1}$-loss as follows,
\begin{equation}
\label{equ:reconstruction_loss}
    \mathcal{L}_{sym}= \left\|P^{'}-\varepsilon (P,r,t,r_{gt},t_{gt}) \right\|_{1}
\end{equation}
where $P,P^{'}$ denotes input point cloud and corresponding point cloud, respectively. $\varepsilon(*)$ depends on the symmetry type. 
So far, the overall loss function can be states as follows,
\begin{equation}
\label{equ:overall_loss}
    \mathcal{L}_{all}= \mathcal{L}_{mm}^{NCE} + \mathcal{L}_{(img,text)}^{CE} + \mathcal{L}_{pose} + \mathcal{L}_{sym} 
\end{equation}

\begin{figure*}[htbp]
    \includegraphics[width=\textwidth]{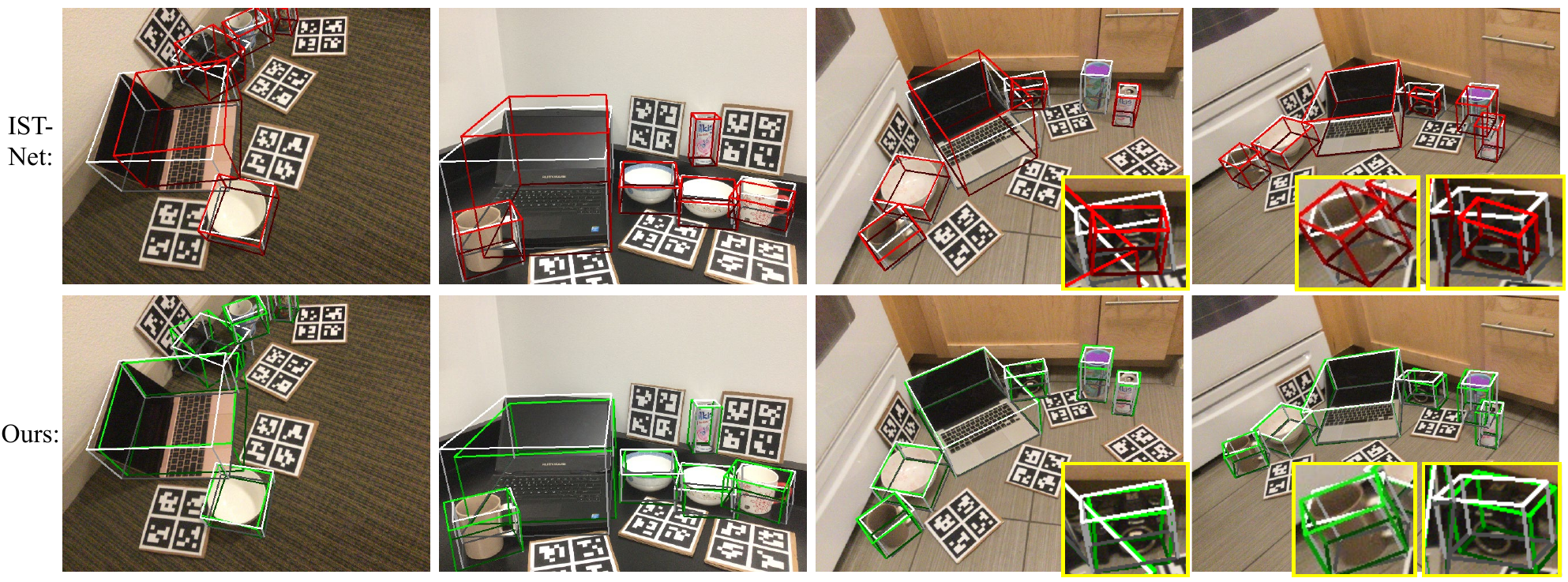}
    \caption{\textbf{Qualitative results of the IST-Net \textcolor{red}{(red line)} and our method \textcolor{green}{(green line)}.} The ground truth results are shown with white lines. The detail comparison areas are annotated with yellow lines. 
    }\label{fig:vis_real}
    \vspace{-0.3cm} 
\end{figure*}

\section{Experiments}
\label{sec:Experiments}
\textbf{Datasets.} We evaluate CLIPose on two mainstream benchmarks\cite{wang2019normalized} for category-level object pose estimation. 
CAMERA25 is a synthetic dataset that provides 300K synthetic RGB-D images of objects rendered on virtual background, with 25K images are withhold for testing. It contains 6 categories of objects, \ie \emph{bottle, bowl, camera, can, laptop} and \emph{mug}. 
REAL275 is a more challenge real-world dataset that contains the same categories as CAMERA25. The training data consists of 4.3K images from 7 scenes, while testing data includes 2.75K from 6 scenes and 3 objects from each category.

{ \bf Evaluation Metrics.} We report the mean average precision (mAP) in $n$° and $m$ cm to evaluate the accuracy of object pose estimation following NOCS~\cite{wang2019normalized}.
The $n$ and $m$ indicate the prediction error of rotation and translation, respectively.
We leverage the combination of rotation and translation metrics of 5°2\emph{cm}, 5°5\emph{cm}, 10°2\emph{cm}, 10°5\emph{cm} and 10°10\emph{cm}, which means the estimation is considered correct when the error is below a threshold. 
Similar to NOCS, for symmetry objects (\emph{bottles, bowls, and cans}), we ignore the rotation error around the symmetry axis.

\begin{table}[htbp]
    \small
    \centering
    \setlength\tabcolsep{4pt}
    \renewcommand\arraystretch{1.2} 
    \caption{\textbf{Comparisons with state-of-the-art methods on CAMERA25 dataset.} Overall best results are in bold and the second best results are underlined. \textbf{Prior} refers to whether the method necessitates prior category information. 
    ‘-’ indicates no results reported under this metric in original paper.
    ($\uparrow$) represents a higher value indicating better performance.}
    \begin{tabular}{r|c|cccc}
    \toprule[1.2pt]
   Methods & Prior & 5°2\emph{cm}$\uparrow$ & 5°5\emph{cm}$\uparrow$ & 10°2\emph{cm}$\uparrow$  & 10°5\emph{cm}$\uparrow$ \\
    \midrule
    NOCS\cite{wang2019normalized}        & $\times$    & 32.3      &40.9          & 48.2    &64.6     \\
    DualPoseNet\cite{lin2021dualposenet}      &$\times$     &29.3        & 35.9           &50.0    &66.8    \\
    GPV-Pose\cite{zheng2023hs}                & $\times $    & 72.1      & 79.1          & -    & 89.0     \\
    IST-Net\cite{liu2023prior}                   & $\times $   & 71.3       & 79.9          & 79.4   &{\ul89.9}     \\
    HS-Pose\cite{zheng2023hs}                         & $\times $    & 73.3      & {\ul80.5}          & 80.4    & 89.4     \\
    \hline
    SPD\cite{tian2020shape}        & \checkmark    & 54.3      &59.0          & 73.3    &81.5     \\
    SAR-Net\cite{lin2022sar}                   & \checkmark    & 66.7       & 70.9          & 75.3   &80.3     \\
    SGPA\cite{chen2021sgpa}                                  &\checkmark     &70.7        & 74.5           &\textbf{82.7}    &88.4    \\
    GPT-COPE\cite{zou2023gpt}                                  &\checkmark     &70.4        & 76.5           &81.3    &88.7    \\
    RBP-Pose\cite{zhang2022rbp}                   & \checkmark   & {\ul73.5}       & 79.6          & {\ul82.1}   &89.5    \\

    \midrule
    Ours                             & $\times$    & \textbf{74.8}     & \textbf{82.2}         & 82.0    &  \textbf{91.2}     \\
    \bottomrule[1.2pt]
    \end{tabular}
    \label{tab:compare_sota_camera}
\end{table}

{ \bf Implementation Details.} 
We follow \cite{tian2020shape,di2022gpv} to generate segmentation masks with the off-the-shelf object detector, Mask-RCNN\cite{he2017mask} and resize them to $224 \times 224$.
We uniformly sample 1024 points to form the object point clouds and feed them as input to 3D encoder, whose backbone is optional according to the needs of the task. We select the modified 3D-GCN in \cite{zheng2023hs} as the encoder. 
The output of the 3D encoder is a concatenation of the outputs of 3D-GCN layers, whose shape is $[1024,1280]$. The point cloud features used for contrastive learning are the output of final 3D-GCN layer after projection and pooling, whose shape is $[1,512]$.
We fix the temperature parameter $\tau$ and set it as 0.07 similar to \cite{radford2021learning}.
We adopt several commonly used data augmentation in category-level object pose estimation tasks, including random uniform noise, random rotational and translational perturbations, and bounding box-based adjustment similar to \cite{chen2021fs,di2022gpv}.
For a fair comparison, the loss terms and their parameters related to point reconstruction and pose estimation are kept the same as \cite{di2022gpv}.
We employ the Ranger optimizer and all experiments are conducted on a single RTX3090 GPU with batch size of 24 and base learning rate of 1e-4, and the learning rate is annealed at 72\% of the training phase using a cosine schedule.

\subsection{Comparison with State-of-the-Art Methods}
\label{sec:compare_sota}
{ \bf Performance on REAL275 dataset.}
In \cref{tab:compare_sota}, we present the results of our CLIPose with state-of-the-art methods on REAL275~\cite{wang2019normalized} dataset, highlighting a significant advancement in performance.
We deal with all categories with a single model.

As shown in \cref{tab:compare_sota}, our method outperforms the state-of-the-art performance in all metrics. 
As for the current state-of-the-art method, HS-Pose~\cite{zheng2023hs}, we still perform significant improvements in most of the metrics, \eg, we achieve \textbf{48.7\%} on 5°2\emph{cm}, \textbf{58.3\%} on 5°5\emph{cm} and \textbf{70.4\%} on 10°2\emph{cm} metrics, which outperform HS-Pose~\cite{zheng2023hs} by 3.4\%, 3.4\% and 1.8\%. 
Noteworthy, we achieve results on the REAL275 dataset, surpassing the impressive thresholds of 85\% and 86\% in the 10°5\emph{cm} and 10°10\emph{cm} metrics, respectively. These exceptional outcomes further support the efficacy of our approach.

Our approach solely relies on depth data as input and does not require a prior models. Even when compared with the methods that take RGB-D data and category prior as input, our method is still superior on all metrics. 
Especially, we achieve \textbf{85.2\%} on 10°5\emph{cm} and \textbf{86.2\%} on 10°10\emph{cm}, which surpass the state-of-the-art method, IST-Net~\cite{liu2023prior}, with a large gap by 5.6\% and 4.6\%.
In addition, we present a qualitative comparison between CLIPose and IST-Net~\cite{liu2023prior}, as shown in \cref{fig:vis_real}.
Our method achieves better size and pose estimation (\eg the laptops in the first three columns), and handles complex and small shapes better (\eg the cameras and mugs in the last two columns).

\begin{table*}[htbp]
    \small
    \centering
    \setlength\tabcolsep{3.8pt}
    \renewcommand\arraystretch{1.2} 
    \caption{\textbf{Ablation studies on REAL275 dataset.} Overall best results are in bold and the second best results are underlined. ($\uparrow$) represents a higher value indicating better performance.
    The default length of prompt tokens is 10.
    $D_0$ indicates the full version of the network. }
    \begin{tabular}{c|cc|cc|ccc|ccccc}
    \toprule[1.2pt]
     \multirow{3}{*}{ID} & \multicolumn{2}{c|}{Contrastive Loss (PC)} & \multicolumn{2}{c|}{Fine-Tune CLIP}& \multicolumn{3}{c|}{Classification Learning}  & \multicolumn{5}{c}{mAP$\uparrow$} \\ 
      \cline{2-13}
     ~ & $\mathcal{L}_{(PC,img)}^{NCE}$ & $\mathcal{L}_{(PC,text)}^{NCE}$  & \doubleline{Text}{with Pose} & \doubleline{Prompt}{Tokens} & $\mathcal{L}_{(img,text)}^{CE}$& $\mathcal{L}_{(img,text)}^{NCE}$ & one-hot   & 5°2\emph{cm} & 5°5\emph{cm} & 10°2\emph{cm}  & 10°5\emph{cm}& 10°10\emph{cm}\\ 
    \midrule
     $A_0$ &  &  &  &  &  &   &    & 43.8   & 54.5 & 65.4  & 81.8 & 82.6 \\
     \midrule
     $B_0$ & \checkmark &  &  &  &  &  &   & 45.3   & 54.6 & 67.7 & 83.2 &83.4 \\
    $B_1$ &  & \checkmark &  &  & &  &  & 42.4   & 53.2 & 65.5 & 82.2 &84.0 \\
    $B_2$ & \checkmark &\checkmark &  &  & &   & & 45.9   & 55.5 & 68.3& 83.5 &84.4 \\
    \midrule
    $C_0$ & \checkmark & \checkmark & \checkmark &  & &  &   & 46.2   & 56.8 & 69.4 &  83.3  &84.5 \\
    $C_1$ & \checkmark & \checkmark &    & \checkmark& &  &   & 47.5   & 57.1 & 69.6 &  {\ul84.5}  &{\ul85.9}\\
    $C_2$ & \checkmark & \checkmark & \checkmark & \checkmark &  & &   & {\ul48.5} & {\ul57.5} & {\ul70.3} & 84.2 &85.6  \\
    \midrule
    \rowcolor{mygray}
    $D_0$ & \checkmark & \checkmark &\checkmark & \checkmark & \checkmark& &      &\textbf{48.7} & \textbf{58.2} & \textbf{70.4} & \textbf{85.2} & \textbf{86.2}\\
    $D_1$ & \checkmark & \checkmark &\checkmark & \checkmark & & \checkmark     & 
     &44.1 & 52.5 & 66.3 & 80.8 & 84.5 \\
     $D_2$ & \checkmark & \checkmark &\checkmark & \checkmark & &      & \checkmark & 47.3         &57.1        & 69.8         & 84.0  &84.8\\
    \bottomrule[1.2pt]
    \end{tabular}
    \label{tab:ablation_study}
    \vspace{-0.1cm} 
\end{table*}
\begin{table}[htbp]
    \small
    \centering
    \setlength\tabcolsep{1.3pt}
    \renewcommand\arraystretch{1.2} 
    \caption{\textbf{Ablation studies on prompt length and the insertion of prompt tokens.} Overall best results are in bold and the second best results are underlined. ($\uparrow$) represents a higher value indicating better performance. Other network settings remain default.
    }
    \begin{tabular}{c|c|c|ccccc}
    \toprule[1.2pt]
     \multirow{2}{*}{ID} & \multirow{2}{*}{\doubleline{Length}{of Tokens}} & \multirow{2}{*}{\doubleline{Location}{of Insertion}}  & \multicolumn{5}{c}{mAP$\uparrow$} \\ 
      \cline{4-8}
     ~ & ~ & ~  & 5°2\emph{cm} & 5°5\emph{cm} & 10°2\emph{cm}  & 10°5\emph{cm} & 10°10\emph{cm}\\ 
     \midrule
     0 & \multirow{2}{*}{10} & Prepend   & \textbf{48.7}  & {\ul58.2} & \textbf{70.4}  & \textbf{85.2} & \textbf{86.2}\\
     1 & ~ & Append  & 46.2   & 55.8 & 69.2  & 83.5 & 84.9\\
    \midrule
     2 & 5 &  \multirow{5}{*}{Prepend}  & {\ul48.3}  & \textbf{58.8} & 68.9 & {\ul84.4} &85.1 \\
     3 & 20 & ~  & 47.3   & 56.1 & 69.9 & 83.7 &84.9 \\
     4 & 50 & ~ & 47.8   & 57.0 & {\ul70.2} & 84.1 &{\ul85.3}  \\
      5 & 100 & ~  & 44.3   & 55.5 & 64.1 & 83.0 &84.5 \\
     6 & 200 & ~ & 40.2   & 53.1 & 61.8 & 82.7 & 84.6 \\
    \bottomrule[1.2pt]
    \end{tabular}
    \label{tab:suppl_ablation_study_length_location}
    \vspace{-0.2cm} 
\end{table}

\pgfplotsset{width=\columnwidth,compat=newest}
\begin{figure}[htbp]
\begin{tikzpicture}[scale=0.7]
    \begin{loglogaxis}[
        xlabel={Length of Tokens},
        ylabel={mAP@5°2\emph{cm}},
         ytick={39,41,43,45,47,49,51},
         yticklabels={39,41,43,45,47,49,51},
         width=5in,
         height=2.6in,
        legend pos=south west,
        nodes near coords={$\pgfmathprintnumber{\pgfplotspointmeta}$},
        point meta=explicit symbolic,
         every node near coord/.style={color=black},
        ymajorgrids=true,
        grid style=dashed,
    ]
    \addplot[very thick,color=red,mark=x,mark options={scale=2.0, line width=2.6}]
        coordinates {
        (5,48.3)[5]
        (10,48.7)[10]
        (20,47.3)[20]
        (50,47.8)[50]
        (100,44.3)[100]
        (200,40.2)[200]
        };
        \legend{Performance of different lengths in 5°2cm}
    \end{loglogaxis}
\end{tikzpicture}
\caption{\textbf{Visualization of ablation studies on prompt length.} The model exhibits improved performance when the length is kept at 50 or below. As the length increases, the outcomes tend to decline.}
\label{fig:prompt_length_vis}
\vspace{-0.4cm}
\end{figure}
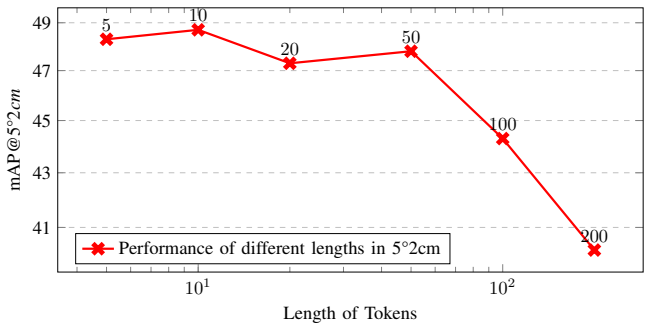


{ \bf Performance on CAMERA25 dataset.}
The results comparison of our method and the state-of-the-art is presented in \cref{tab:compare_sota_camera}. 
From the observation of results, it can be deduced that CLIPose demonstrates superior performance across all metrics except for 10°2\emph{cm}, where our method also achieves highly competitive results, closely approaching the best performance. 
As for the comparison with prior-free methods, our method achieves a remarkable improvement over the current state-of-the-art method, HS-Pose~\cite{zheng2023hs}, exceeding it by nearly 2\% in both the rigorous 5°2\emph{cm} and 5°5\emph{cm} evaluation metrics.
When compared to methods that take category prior as input, our method still maintains a notable advantage without relying on any additional information.
This indicates that our method has a strong ability to comprehensively estimate rotation, translation, and size.
It is worth noting that CAMERA25 is a synthetic dataset that contains many other extraneous objects, our method can still achieve the state-of-the-art performance.
This superior performance on synthetic dataset further proves the effectiveness of our method. 


\subsection{Ablation Study}
\label{sec:ablation_study}
To validate the proposed contributions, we meticulously conduct ablation studies on the REAL275\cite{wang2019normalized} dataset.
We divide the contribution terms into three aspects and study their influences through different combinations.
The full ablation study results are shown in \cref{tab:ablation_study}. $A_0$ indicates the baseline method that only utilize point reconstruction and pose estimation losses.

{ \bf 1) Effect of Contrastive Loss.}
The contrastive learning architecture is the key ingredient of our method, which includes loss functions between the point cloud and the image $\mathcal{L}_{(PC,img)}^{NCE}$ as well as the point cloud and the text $\mathcal{L}_{(PC,text)}^{NCE}$.
To demonstrate the effectiveness of abundant semantics captured in image and text modalities, we train our network utilizing the single contrastive learning loss function, which in practice can be thought as aligning the point cloud and image or text features individually. 
The results in \cref{tab:ablation_study}, specifically the [$B_0$] and [$B_1$] rows, indicate that aligning the representations of two modalities, either point cloud and image or point cloud and text, positively influences the baseline method. 
By observing the results, we found that aligning point cloud and image can improve the performance in 5°2\emph{cm} to 10°10\emph{cm} metrics.
As shown in [$B_2$] row, when we incorporate the two losses, \ie, perform a complete multi-modal contrastive learning, the performance in all metrics would be improved.
\begin{table}[htbp]
    \small
    \centering
    \setlength\tabcolsep{1.8pt}
    \renewcommand\arraystretch{1.2} 
    \caption{\textbf{Ablation studies on the involved transformer layers of image encoder and joint training with text encoder.} Overall best results are in bold and the second best results are underlined. 
    ($\uparrow$) represents a higher value indicating better performance.
    Other network settings remain default.}
    \begin{tabular}{c|c|c|ccccc}
    \toprule[1.2pt]
     \multirow{2}{*}{ID} & \multirow{2}{*}{\doubleline{Transformer}{Layers}} & \multirow{2}{*}{\doubleline{Joint}{Training}}  & \multicolumn{5}{c}{mAP$\uparrow$} \\ 
      \cline{4-8}
     ~ & ~ & ~   & 5°2\emph{cm} & 5°5\emph{cm} & 10°2\emph{cm}  & 10°5\emph{cm} & 10°10\emph{cm}\\ 
     \midrule
     0 & \multirow{2}{*}{1 $\to$ 1} & -  & \textbf{48.7}  & \textbf{58.2} & \textbf{70.4}  & \textbf{85.2} & \textbf{86.2}\\
     1 & ~ & \checkmark  & {\ul47.0}   & 56.6 & 68.6  & 83.4 & 84.5\\
    \midrule
     2 & 1 $\to$ 3 &  -   & 45.4   &53.5 & 68.4 & 82.7 &84.0 \\
     3 & 1 $\to$ 6  & -  & 46.9   & 56.8 & 69.3 & {\ul84.7} &{\ul85.9} \\
     4 & 1 $\to$ 12  & -  & 45.9   & 54.7 & {\ul69.9} & 83.6 &85.0 \\
      5 & 6 $\to$ 12  & - & {\ul47.0}   & {\ul57.3} & 69.5  & 83.6 & 84.7 \\
     6 & 9 $\to$ 12  & -  & 45.7   & 55.2 & 67.9 & 82.7&84.1 \\
      7 & 12 $\to$ 12  & -  & 46.2   & 56.7 & 67.8 & 83.0&84.1 \\
    \bottomrule[1.2pt]
    \end{tabular}
    \label{tab:suppl_ablation_study_layer_joint}
    \vspace{-0.3cm}
\end{table}
We can achieve comparable results to HS-Pose\cite{zheng2023hs} without fine-tuning the CLIP model, which proves the effectiveness of contrastive learning.

{ \bf 2) Effect of Fine-Tune Manner.}
The fine-tuning of the CLIP model includes both text and image portions.
The category loss of network, the text description with pose information and the additional prompt tokens are employed to fine-tune the CLIP model to make it more suitable for the pose estimation task.
To verify the benefit of two contributions, we incorporate each of them on top of the complete contrastive learning, as shown in [$C_0$] and [$C_1$] rows.
It is clear that two combinations all boost the performance. 
Specifically, the modified text description improves the performance on some metrics, \eg 5°2\emph{cm}, 5°5\emph{cm} and 10°2\emph{cm}.
Furthermore, the prompt tokens for image encoder achieve better results on all metrics. 
The performance of this version ([$C_1$]) in 10°5\emph{cm} (\textbf{84.5\%}) and 10°10\emph{cm} (\textbf{85.9\%}), which exhibits that the additional prompt tokens have a significant improvement in network performance.
Eventually, we combine two manners together for evaluation, as shown in [$C_2$]. This version achieves remarkable results, closely following the performance of the best version [$E_0$].

The length of prompt tokens is also a factor influencing the results. To achieve better fine-tuning results, we perform further ablation studies on prompt tokens. 
We specifically investigate several aspects, including the length of prompt tokens, the location of token insertion, the number of transformer layers involved, and the joint fine-tuning training with the text encoder. 
All ablation results are presented in \cref{tab:suppl_ablation_study_length_location} and \cref{tab:suppl_ablation_study_layer_joint}. 
Note that our default settings are as follows: the length of prompt tokens is 10, insertion of tokens at the beginning (prepend), and involvement of the first transformer layer (total is 12). 

From \cref{tab:suppl_ablation_study_length_location}, we evaluate the performance \wrt different prompt lengths, the length of the tokens is set to 5, 10, 20, 50, 100, and 200 and compare their performance. 
As can be observed, the default setting (length is 10) achieves the best result. As shown in ID-2 row, competitive results are also obtained when the length is set to 5.
ID-3 and ID-4 rows exhibit increasing the length of tokens within certain limits can also maintain the performance.
The rows ID-5 and ID-6 demonstrate that when the length of the token is close or equivalent to the default length of the image embedding (196), it results in a performance decrease. 
The \cref{fig:prompt_length_vis} intuitively illustrates the impact of prompt length on performance.
Additionally, we keep a default length of tokens and compare the performance between insertion at the front (prepend) and end (append) of the image embedding, as shown in rows ID-0 and ID-1. 
The overall performance shows that the difference between these two ways is not significant, with token insertion in the prepend position slightly outperforming the append position.

In \cref{tab:suppl_ablation_study_layer_joint}, we further evaluate the performance when changing the transformer layers of image encoder, \eg $1\to 3$ means fine-tuning in the first three layers, and $9\to 12$ denotes fine-tuning in the last three layers. 
In the Transformer encoder, the layers closer to the input primarily focus on learning the local structures of the input embedding, while the layers towards the end tend to specialize in capturing global features and high-level abstractions of the embedding.
The default setting, \ie inserting tokens in the first layer still surpasses other settings.
The overall performance exhibits an interesting result, with the performance being better when the number of layers added is 6 compared to the cases with 3 and 12 layers, while adding tokens in the later layers achieves superior results. 
We argue that introducing prompt tokens in both the first and later transformer layers does not significantly impact the extraction of image features while allowing for effective fine-tuning of the image encoder. 
Moreover, we validated the joint training of image and text encoders, assessing their performance when fine-tuned simultaneously. 
As can be observed in rows ID-0 and ID-1, the results of joint training exhibits a slight decline.
This further demonstrates the speculation in \cref{sec:prompt_tune} that the incorporation of pose parameters and text description has already provided the pose-sensitive information, and does not require fine-tuning of text encoder. 

{ \bf 3) Ablate on Classification Learning.}
In \cref{sec:aligning}, we have discussed that we construct the multi-categorical Contrastive Estimation (CE) loss between image and text modalities to introduce object classification information and further enhance the model’s learning of category features. 
To validate the assumptions we presented in \cref{sec:aligning}, we compare the results of applying different losses between image and text modalities.
As shown in \cref{tab:ablation_study} [$D_0$] rows, the CE loss can effectively assist the network to achieve the best performance. 
This version donates that we are leveraging CLIP's inherent categorization ability, and we treat [$D_0$] as the final version of our method.
On the contrary, applying NCE loss ([$D_1$]) means aligning the image and text representations that are already aligned in the CLIP model. 
The results of [$D_1$] has a large gap compared to [$D_0$], with some metrics even inferior to the baseline ([$A_0$]). 
We speculate that the text and image re-alignment does not directly facilitate the feature learning of the 3D encoder. 
Additionally, since we have fine-tuned the CLIP's image encoder, the re-alignment between text and image may even impact the performance of the CLIP model itself. 
\begin{table*}[htbp]
    \small
    \centering
    \setlength\tabcolsep{7.5pt}
    \renewcommand\arraystretch{1.2} 
    \caption{\textbf{Ablation studies on feature fusion.} ID-0 denotes baseline method and is colored in \colorbox{gray!30}{gray}. [\textbf{Major}] term of fusion manner represents the core concepts of fusion and [\textbf{Specific}] term indicates the manner of detailed. \textbf{GP} denotes global pooling and \textbf{PC} represents Point Cloud. Other network settings remain default. 
    (\textcolor{red}{$_\downarrow$}) indicates the difference that has decreased compared to the baseline ID-0.
    ($\uparrow$) represents a higher value indicating better performance. 
    }
    \begin{tabular}{c|c|c|cc|ccccc}
    \toprule[1.2pt]
     \multirow{3}{*}{ID} & \multicolumn{2}{c|}{Fusion Manner} & \multicolumn{2}{c|}{Feature for Fusion}  & \multicolumn{5}{c}{mAP$\uparrow$} \\ 
      \cline{2-10}
     ~ & Major & Specific  &\doubleline{Point Cloud}{\& Image}  & \doubleline{Point Cloud}{\& Text} & 5°2\emph{cm} & 5°5\emph{cm} & 10°2\emph{cm}  & 10°5\emph{cm} & 10°10\emph{cm}\\ 
     \midrule
    \rowcolor{mygray}
     0 & - & - & - & -    & 43.8   & 54.5 & 65.4  & 81.8 & 82.6\\
    \midrule
     1 & \multirow{2}{*}{Concat} & Before GP & \checkmark & -     & 41.4$_{\textcolor{red}{\downarrow 2.4}}$   & 51.7$_{\textcolor{red}{\downarrow 2.8}}$ & 61.7$_{\textcolor{red}{\downarrow 3.7}}$  & 79.2$_{\textcolor{red}{\downarrow 2.6}}$ & 80.1$_{\textcolor{red}{\downarrow 2.5}}$\\
     \cline{1-1}
     \cline{3-3}
     2 & ~ & After GP & - &  \checkmark     & 42.2$_{\textcolor{red}{\downarrow 1.6}}$   & 51.9$_{\textcolor{red}{\downarrow 2.6}}$ & 63.9$_{\textcolor{red}{\downarrow 1.5}}$ & 80.4$_{\textcolor{red}{\downarrow 1.4}}$ &81.2$_{\textcolor{red}{\downarrow 1.4}}$ \\
    \midrule
     3 & \multirow{4}{*}{\doubleline{Cross-}{Attention}} & \multirow{2}{*}{PC as Q} & \checkmark & -   &   0.1$_{\textcolor{red}{\downarrow 43.7}}$   &  0.4$_{\textcolor{red}{\downarrow 54.1}}$ &  2.3$_{\textcolor{red}{\downarrow 63.1}}$ &  5.1$_{\textcolor{red}{\downarrow 76.7}}$&  5.3$_{\textcolor{red}{\downarrow 77.3}}$ \\
     \cline{1-1}
     4 & ~ & ~ & - & \checkmark   &  0.8$_{\textcolor{red}{\downarrow 43.0}}$ &  1.1$_{\textcolor{red}{\downarrow 53.4}}$ &  5.7$_{\textcolor{red}{\downarrow 59.7}}$ & 8.1$_{\textcolor{red}{\downarrow 73.7}}$ &  8.2$_{\textcolor{red}{\downarrow 74.4}}$   \\
      \cline{1-1}
     \cline{3-3}
      5 & ~ & \multirow{2}{*}{PC as KV} & \checkmark & -  & 5.1$_{\textcolor{red}{\downarrow 38.7}}$   & 14.6$_{\textcolor{red}{\downarrow 39.9}}$ & 18.5$_{\textcolor{red}{\downarrow 46.9}}$  & 46.6$_{\textcolor{red}{\downarrow 35.2}}$ & 47.6$_{\textcolor{red}{\downarrow 35.0}}$ \\
     \cline{1-1}
     6 & ~ & ~ & - &\checkmark  & 10.2$_{\textcolor{red}{\downarrow 33.6}}$   & 13.6$_{\textcolor{red}{\downarrow 40.9}}$ & 29.0$_{\textcolor{red}{\downarrow 36.4}}$ & 44.4$_{\textcolor{red}{\downarrow 37.4}}$&46.6$_{\textcolor{red}{\downarrow 36.0}}$ \\
       \midrule
     7 & \multirow{2}{*}{\doubleline{Transformer}{Encoder}} & Plus(regular) & \checkmark& - & 38.3$_{\textcolor{red}{\downarrow 5.5}}$    & 47.3$_{\textcolor{red}{\downarrow 7.2}}$   & 60.2$_{\textcolor{red}{\downarrow 5.2}}$ & 73.4$_{\textcolor{red}{\downarrow 8.4}}$ & 75.4$_{\textcolor{red}{\downarrow 7.2}}$  \\
     \cline{1-1}
     \cline{3-3}
     8 & ~ & Concat & - & \checkmark  & 24.4$_{\textcolor{red}{\downarrow 19.4}}$   & 36.6$_{\textcolor{red}{\downarrow 17.9}}$ & 45.1$_{\textcolor{red}{\downarrow 20.3}}$ & 67.1$_{\textcolor{red}{\downarrow 14.7}}$ &68.7$_{\textcolor{red}{\downarrow 13.9}}$ \\
    \bottomrule[1.2pt]
    \end{tabular}
    \label{tab:suppl_ablation_study_fuse}
\end{table*}
\begin{table}[htbp]
    \small
    \centering
    \setlength\tabcolsep{6pt}
    \renewcommand\arraystretch{1.2} 
    \caption{\textbf{Comparisons of using different 3D Encoder.} CLIPose adopts the 3D Encoders from GPV-Pose~\cite{di2022gpv} or HS-Pose~\cite{zheng2023hs} and performs a comparison with these two methods. ($\uparrow$) represents a higher value indicating better performance.}
    \begin{tabular}{r|ccccc}
    \toprule[1.2pt]
   Methods  & 5°2\emph{cm}$\uparrow$ & 5°5\emph{cm}$\uparrow$ & 10°2\emph{cm}$\uparrow$  & 10°5\emph{cm}$\uparrow$ \\
    \midrule
    GPV-Pose\cite{di2022gpv}                          & 32.5 & 43.3         & 58.2 & 76.6             \\
    Ours(3D-GCN)       & 42.9     &54.4         & 63.7    &81.5     \\
     \midrule
    HS-Pose\cite{zheng2023hs}                   & 45.3          &54.9         & 68.6         & 83.6     \\
    Ours(HS-Layer)                       & 48.7         & 58.3 & 70.4 & 85.2    \\
    \bottomrule[1.2pt]
    \end{tabular}
    \label{tab:suppl_3d_encoder}
    \vspace{-0.3cm} 
\end{table}

Actually, there are several intuitive ways to incorporate object classification information, such as using categorical one-hot vectors~\cite{di2022gpv,zheng2023hs}.
We further perform ablation studies to investigate whether our method could outperforms explicit one-hot vectors modeling of object categories, as shown in [$D_2$].
It can be clearly observed that the performance of the model is not significantly different from [$C_2$] after the introduction of categorical one-hot vectors, with a slight decrease in the results for some metrics.
We suppose that during the multimodal learning process, the category information carried by the one-hot vectors may mix with the spatial information of the point cloud, preventing the network from effectively learning valuable classification information.

{ \bf 4) Ablate on 3D Encoder.}
CLIPose can leverage any 3D encoder to extract point cloud features. 
The default setting of CLIPose's 3D encoder is the HS-Layer from HS-Pose~\cite{zheng2023hs}.
We also adopt the 3D-GCN~\cite{lin2020convolution} similar to GPV-Pose~\cite{di2022gpv} to evaluate the performance.
The overall results are shown in \cref{tab:suppl_3d_encoder}, as can be observed that different 3D encoders can give varying performance outcomes. 
What's more, under the condition of identical 3D encoders, our method demonstrates superior performance compared to the original methods.

Specifically, our method surpasses GPV-Pose~\cite{di2022gpv} by more than 10\% in rigorous 5°2\emph{cm} and 5°5\emph{cm} metrics. Moreover, it exceeds HS-Pose~\cite{zheng2023hs} by 3.4\% and 3.4\% in the same two metrics, which further proves the effectiveness of our method.

\begin{table}[htbp]
    \small
    \centering
    \setlength\tabcolsep{4pt}
    \renewcommand\arraystretch{1.2} 
    \caption{\textbf{Ablation studies on different temperature parameters $\tau$.} Overall best results are in bold and the second best results are underlined. ($\uparrow$) represents a higher value indicating better performance.}
    \begin{tabular}{c|c|c|ccccc}
    \toprule[1.2pt]
   ID & $\tau$ &logit\_scale  & 5°2\emph{cm}$\uparrow$ & 5°5\emph{cm}$\uparrow$ & 10°2\emph{cm}$\uparrow$  & 10°5\emph{cm}$\uparrow$ \\
    \midrule
    0 &0.2            &    5.0   &    \textbf{48.8}  & \textbf{58.9}      & {\ul69.4} & {\ul83.8}             \\
     \rowcolor{mygray}
    1&0.07     &   14.3       & {\ul48.7}          &{\ul58.3}         & \textbf{70.4}         & \textbf{85.2}    \\
    2&0.03    &       33.3             & 46.3         & 55.9 & 68.2 & 83.6    \\
    3&0.02    &       50.0             & 46.8        & 56.5 & 67.2 & 81.8    \\
    4&0.001    &       100.0             & 45.7         & 54.6 & 67.4 & 81.4    \\
    \bottomrule[1.2pt]
    \end{tabular}
    \label{tab:suppl_temperature}
    \vspace{-0.3cm} 
\end{table}

{ \bf 5) Ablate on Temperature Parameter.}
The temperature $\tau$ is a vital hyper-parameter in contrastive learning, as it influences model's discriminative ability towards negative samples.
According to \cref{equ:multimodal_loss}, temperature parameter $\tau$ will transform into the form of coefficient logit\_scale as follows,
\begin{equation}
\label{equ:sullp_scale}
     logit\_scale = log \ e^{\left ( 1/\tau \right ) } =1/\tau
\end{equation}
The performance comparison of the different temperature parameters is shown in \cref{tab:suppl_temperature}. 
As can be observed, if $\tau$ is set too small, the data distribution will become more concentrated, leading to the decline in model's generalization performance. 
Conversely, if we gradually increase $\tau$, the distribution becomes smoother, which is more conducive to the model's learning process.

{ \bf 6) Ablate on Feature Fusion.}
We ultimately utilize point cloud features for pose estimation.
In practice, we also attempt to fuse image or text features into point cloud features and then use the fused features for training. 
Intuitively, the fused features contains more information and can achieve better performance. 
We separately try several fusion manners among multi modalities, \eg direct concatenation, single cross-attention layer and modified transformer encoder, to fuse the different features. 
However, the performance of the network is \textcolor{red}{\textbf{not}} improved, but rather drops compared with the baseline method. 

\begin{table}[htbp]
    \small
    \centering
    \setlength\tabcolsep{6pt}
    \renewcommand\arraystretch{1.2} 
    \caption{\textbf{Ablation studies on the format of classification learning and pose information in textual descriptions.} ID-0 denotes full version model the same as [$D_0$] in \cref{tab:ablation_study} and is colored in \colorbox{mygray}{gray}. Other network settings remain default. ($\uparrow$) represents a higher value indicating better performance. }
    \begin{tabular}{c|ccc|cccc}
    \toprule[1.2pt]
     \multirow{2}{*}{ID} & \multicolumn{3}{c|}{Pose Information}  & \multicolumn{4}{c}{mAP$\uparrow$} \\ 
      \cline{2-8}
     ~ &\emph{w} & \emph{w/o} & \emph{rand} & 5°2\emph{cm} & 5°5\emph{cm} & 10°2\emph{cm}  & 10°5\emph{cm} \\ 
     \midrule
    \rowcolor{mygray}
     0 & \checkmark & -    & -   &    48.7  & 58.2      & 70.4 & 85.2\\
     1 & - & \checkmark     & -   & 47.3         &56.8        & 69.8        & 85.2\\
     2  & - &  -     & \checkmark  &  47.7   & 57.0  & 68.3  & 83.3  \\
    \bottomrule[1.2pt]
    \end{tabular}
    \label{tab:suppl_ablation_textpose}
\end{table}

In \cref{tab:suppl_ablation_study_fuse}, 
for the direct concatenation, we attempt two fusion ways according to the steps of fusion and global pooling. 
If we fuse features before global pooling, we replicate the pooled image/text features to the number of sampled point clouds, then concatenate the image/text features after point cloud features.
If we fuse features after global pooling, we project the pooled different features to the same dimension and then fuse to estimate the pose. 
From the results, we can observe that a slight decrease in performance. 
As for fusion using cross-attention layer, whose principle formula is as follows, 
\begin{equation}
\label{equ:sullp_cross_attn}
     Attention \left (Q, K, V  \right ) = softmax\left (\frac{QK^{T}}{\sqrt{d_k}}  \right )V
\end{equation}
where $d_k$ denotes the dimension of $K$. Typically, one modal features are used as $Q$ and the other as $K$ and $V$. 
We can clearly see that $V$ is the major part of fused features, and the $softmax$ function is use to calculate the weights. 
We take the point cloud features as $Q$ and $KV$ respectively and the results is presented in ID-3 to ID-6. 
The experiments reveal a significant drop in performance when using the point cloud features as $Q$.
When using point cloud features as $KV$, \ie as the main body of the fused features, the results are slightly better than the former but still large lag behind the baseline methods.

Finally, we also try to leverage the Transformer~\cite{vaswani2017attention} encoder since the output features of attention layer are skip connected with $Q$. 
Thus, $Q$ becomes the major part of fused features.
Hence, we introduce the cross-attention layer into the Transformer encoder to replace the original self-attention layer and fuse the features using the modified Transformer encoder, where we take the point cloud features as $Q$. 
The results is shown in ID-7 and ID-8 of \cref{tab:suppl_ablation_study_fuse}. 
Note that we make a further distinguish in modified Transformer encoder, where \textbf{Plus} refers to regular skip connection.
\textbf{Concat} indicates concatenation, the fused features will be projected back to the original dimension. 
From the results, we observe that the performance is better than using a single cross-attention layer but slightly behind the direct concatenation. 
However, all feature fusion manners result in a relative decrease in performance compared to the baseline ID-0. 
Therefore, we directly utilize point cloud features for recovering pose parameters.


{ \bf 7) Ablate on Pose Information in Textual Descriptions.}
Considering that we introduce rotation and translation values in the textual descriptions to make CLIP more adaptable to object's pose information.
It is necessary to validate whether the CLIP text encoder can learn meaningful information from these textual descriptions.
To verify this, we employed three different settings: one with correct pose information denoted as \emph{w}, one without pose information, only containing object category descriptions denoted as \emph{w/o} and one with random pose information denoted as \emph{rand}.
As shown in \cref{tab:suppl_ablation_textpose}, introducing correct pose information (ID-0) proves to be effective in enhancing prediction accuracy, especially on some rigorous metrics (\eg 1.4\% on 5°5\emph{cm}).
On the contrary, when random pose information (ID-2) is introduced, the overall performance declines. 
In most metrics, the results are even inferior compared to the scenario without pose information (ID-1).
This further indicates that guidance from the textual modality can effectively assist in point cloud learning.

\section{Conclusion}
\label{sec:Conclusion}
In this paper, we propose a novel and effective category-level pose estimation network, named CLIPose.
CLIPose employs the pre-trained vision-language model for the category-level pose estimation task.
To fully leverage the abundant semantic knowledge already captured in image and text modalities, CLIPose aligns features from point clouds and two other modalities by means of contrastive learning. 
The alignment operation allows the network learn more robust and representative category-specific information.
Furthermore, CLIPose fine-tunes the pre-trained CLIP model by incorporating the prompt tokens and pose information, which makes CLIP more suitable to the pose estimation, resulting in better estimation performance.
Extensive experiments on the challenge benchmarks show the effectiveness of our method in both efficiency and accuracy.

In the future, we consider exploring two aspects. First, building upon the point cloud data, ewe aim to investigate leveraging RGB data to acquire additional pose information, enabling the correction of poses predicted using point cloud data. 
Second, the impressive capability of the CLIP pre-trained model in zero-shot estimation tasks is impressive. We expect to achieve promising zero-shot estimation performance in pose estimation tasks as well.
We intend to create large-scale point cloud datasets using generative models and further explore the potential of multi-modal contrastive learning in the context of 3D modality.

\bibliographystyle{IEEEtran}
\bibliography{IEEEabrv,myrefs}

\begin{thebibliography}{10}
\providecommand{\url}[1]{#1}
\csname url@samestyle\endcsname
\providecommand{\newblock}{\relax}
\providecommand{\bibinfo}[2]{#2}
\providecommand{\BIBentrySTDinterwordspacing}{\spaceskip=0pt\relax}
\providecommand{\BIBentryALTinterwordstretchfactor}{4}
\providecommand{\BIBentryALTinterwordspacing}{\spaceskip=\fontdimen2\font plus
\BIBentryALTinterwordstretchfactor\fontdimen3\font minus
  \fontdimen4\font\relax}
\providecommand{\BIBforeignlanguage}[2]{{%
\expandafter\ifx\csname l@#1\endcsname\relax
\typeout{** WARNING: IEEEtran.bst: No hyphenation pattern has been}%
\typeout{** loaded for the language `#1'. Using the pattern for}%
\typeout{** the default language instead.}%
\else
\language=\csname l@#1\endcsname
\fi
#2}}
\providecommand{\BIBdecl}{\relax}
\BIBdecl

\bibitem{marchand2015pose}
E.~Marchand, H.~Uchiyama, and F.~Spindler, ``Pose estimation for augmented
  reality: a hands-on survey,'' \emph{IEEE transactions on visualization and
  computer graphics}, vol.~22, no.~12, pp. 2633--2651, 2015.

\bibitem{chen2017multi}
X.~Chen, H.~Ma, J.~Wan, B.~Li, and T.~Xia, ``Multi-view 3d object detection
  network for autonomous driving,'' in \emph{Proceedings of the IEEE conference
  on Computer Vision and Pattern Recognition}, 2017, pp. 1907--1915.

\bibitem{tremblay2018deep}
J.~Tremblay, T.~To, B.~Sundaralingam, Y.~Xiang, D.~Fox, and S.~Birchfield,
  ``Deep object pose estimation for semantic robotic grasping of household
  objects,'' \emph{arXiv preprint arXiv:1809.10790}, 2018.

\bibitem{wang2019densefusion}
C.~Wang, D.~Xu, Y.~Zhu, R.~Mart{\'\i}n-Mart{\'\i}n, C.~Lu, L.~Fei-Fei, and
  S.~Savarese, ``Densefusion: 6d object pose estimation by iterative dense
  fusion,'' in \emph{Proceedings of the IEEE/CVF conference on computer vision
  and pattern recognition}, 2019, pp. 3343--3352.

\bibitem{peng2019pvnet}
S.~Peng, Y.~Liu, Q.~Huang, X.~Zhou, and H.~Bao, ``Pvnet: Pixel-wise voting
  network for 6dof pose estimation,'' in \emph{Proceedings of the IEEE/CVF
  Conference on Computer Vision and Pattern Recognition}, 2019, pp. 4561--4570.

\bibitem{gao2021cloudaae}
G.~Gao, M.~Lauri, X.~Hu, J.~Zhang, and S.~Frintrop, ``Cloudaae: Learning 6d
  object pose regression with on-line data synthesis on point clouds,'' in
  \emph{2021 IEEE International Conference on Robotics and Automation
  (ICRA)}.\hskip 1em plus 0.5em minus 0.4em\relax IEEE, 2021, pp.
  11\,081--11\,087.

\bibitem{zhou2021semi}
G.~Zhou, D.~Wang, Y.~Yan, H.~Chen, and Q.~Chen, ``Semi-supervised 6d object
  pose estimation without using real annotations,'' \emph{IEEE Transactions on
  Circuits and Systems for Video Technology}, vol.~32, no.~8, pp. 5163--5174,
  2021.

\bibitem{he2021ffb6d}
Y.~He, H.~Huang, H.~Fan, Q.~Chen, and J.~Sun, ``Ffb6d: A full flow
  bidirectional fusion network for 6d pose estimation,'' in \emph{Proceedings
  of the IEEE/CVF Conference on Computer Vision and Pattern Recognition}, 2021,
  pp. 3003--3013.

\bibitem{lin2023transpose}
X.~Lin, D.~Wang, G.~Zhou, C.~Liu, and Q.~Chen, ``Transpose: 6d object pose
  estimation with geometry-aware transformer,'' \emph{arXiv preprint
  arXiv:2310.16279}, 2023.

\bibitem{cao2023dgecn++}
T.~Cao, W.~Zhang, Y.~Fu, S.~Zheng, F.~Luo, and C.~Xiao, ``Dgecn++: A
  depth-guided edge convolutional network for end-to-end 6d pose estimation via
  attention mechanism,'' \emph{IEEE Transactions on Circuits and Systems for
  Video Technology}, 2023.

\bibitem{hinterstoisser2011gradient}
S.~Hinterstoisser, C.~Cagniart, S.~Ilic, P.~Sturm, N.~Navab, P.~Fua, and
  V.~Lepetit, ``Gradient response maps for real-time detection of textureless
  objects,'' \emph{IEEE transactions on pattern analysis and machine
  intelligence}, vol.~34, no.~5, pp. 876--888, 2011.

\bibitem{brachmann2016uncertainty}
E.~Brachmann, F.~Michel, A.~Krull, M.~Y. Yang, S.~Gumhold \emph{et~al.},
  ``Uncertainty-driven 6d pose estimation of objects and scenes from a single
  rgb image,'' in \emph{Proceedings of the IEEE conference on computer vision
  and pattern recognition}, 2016, pp. 3364--3372.

\bibitem{xiang2017posecnn}
Y.~Xiang, T.~Schmidt, V.~Narayanan, and D.~Fox, ``Posecnn: A convolutional
  neural network for 6d object pose estimation in cluttered scenes,''
  \emph{arXiv preprint arXiv:1711.00199}, 2017.

\bibitem{wang2019normalized}
H.~Wang, S.~Sridhar, J.~Huang, J.~Valentin, S.~Song, and L.~J. Guibas,
  ``Normalized object coordinate space for category-level 6d object pose and
  size estimation,'' in \emph{Proceedings of the IEEE/CVF Conference on
  Computer Vision and Pattern Recognition}, 2019, pp. 2642--2651.

\bibitem{tian2020shape}
M.~Tian, M.~H. Ang, and G.~H. Lee, ``Shape prior deformation for categorical 6d
  object pose and size estimation,'' in \emph{Computer Vision--ECCV 2020: 16th
  European Conference, Glasgow, UK, August 23--28, 2020, Proceedings, Part XXI
  16}.\hskip 1em plus 0.5em minus 0.4em\relax Springer, 2020, pp. 530--546.

\bibitem{chen2021fs}
W.~Chen, X.~Jia, H.~J. Chang, J.~Duan, L.~Shen, and A.~Leonardis, ``Fs-net:
  Fast shape-based network for category-level 6d object pose estimation with
  decoupled rotation mechanism,'' in \emph{Proceedings of the IEEE/CVF
  Conference on Computer Vision and Pattern Recognition}, 2021, pp. 1581--1590.

\bibitem{di2022gpv}
Y.~Di, R.~Zhang, Z.~Lou, F.~Manhardt, X.~Ji, N.~Navab, and F.~Tombari,
  ``Gpv-pose: Category-level object pose estimation via geometry-guided
  point-wise voting,'' in \emph{Proceedings of the IEEE/CVF Conference on
  Computer Vision and Pattern Recognition}, 2022, pp. 6781--6791.

\bibitem{liu2023prior}
J.~Liu, Y.~Chen, X.~Ye, and X.~Qi, ``Prior-free category-level pose estimation
  with implicit space transformation,'' \emph{arXiv preprint arXiv:2303.13479},
  2023.

\bibitem{zou2023gpt}
L.~Zou, Z.~Huang, N.~Gu, and G.~Wang, ``Gpt-cope: A graph-guided point
  transformer for category-level object pose estimation,'' \emph{IEEE
  Transactions on Circuits and Systems for Video Technology}, 2023.

\bibitem{goyal2021revisiting}
A.~Goyal, H.~Law, B.~Liu, A.~Newell, and J.~Deng, ``Revisiting point cloud
  shape classification with a simple and effective baseline,'' in
  \emph{International Conference on Machine Learning}.\hskip 1em plus 0.5em
  minus 0.4em\relax PMLR, 2021, pp. 3809--3820.

\bibitem{wu20153d}
Z.~Wu, S.~Song, A.~Khosla, F.~Yu, L.~Zhang, X.~Tang, and J.~Xiao, ``3d
  shapenets: A deep representation for volumetric shapes,'' in
  \emph{Proceedings of the IEEE conference on computer vision and pattern
  recognition}, 2015, pp. 1912--1920.

\bibitem{yu2022point}
X.~Yu, L.~Tang, Y.~Rao, T.~Huang, J.~Zhou, and J.~Lu, ``Point-bert:
  Pre-training 3d point cloud transformers with masked point modeling,'' in
  \emph{Proceedings of the IEEE/CVF Conference on Computer Vision and Pattern
  Recognition}, 2022, pp. 19\,313--19\,322.

\bibitem{radford2021learning}
A.~Radford, J.~W. Kim, C.~Hallacy, A.~Ramesh, G.~Goh, S.~Agarwal, G.~Sastry,
  A.~Askell, P.~Mishkin, J.~Clark \emph{et~al.}, ``Learning transferable visual
  models from natural language supervision,'' in \emph{International conference
  on machine learning}.\hskip 1em plus 0.5em minus 0.4em\relax PMLR, 2021, pp.
  8748--8763.

\bibitem{li2021align}
J.~Li, R.~Selvaraju, A.~Gotmare, S.~Joty, C.~Xiong, and S.~C.~H. Hoi, ``Align
  before fuse: Vision and language representation learning with momentum
  distillation,'' \emph{Advances in neural information processing systems},
  vol.~34, pp. 9694--9705, 2021.

\bibitem{yu2022coca}
J.~Yu, Z.~Wang, V.~Vasudevan, L.~Yeung, M.~Seyedhosseini, and Y.~Wu, ``Coca:
  Contrastive captioners are image-text foundation models,'' \emph{arXiv
  preprint arXiv:2205.01917}, 2022.

\bibitem{dang2023instructdet}
R.~Dang, J.~Feng, H.~Zhang, C.~Ge, L.~Song, L.~Gong, C.~Liu, Q.~Chen, F.~Zhu,
  R.~Zhao \emph{et~al.}, ``Instructdet: Diversifying referring object detection
  with generalized instructions,'' \emph{arXiv preprint arXiv:2310.05136},
  2023.

\bibitem{qiu2021vt}
L.~Qiu, R.~Zhang, Z.~Guo, Z.~Zeng, Y.~Li, and G.~Zhang, ``Vt-clip: Enhancing
  vision-language models with visual-guided texts,'' \emph{arXiv preprint
  arXiv:2112.02399}, 2021.

\bibitem{li2022grounded}
L.~H. Li, P.~Zhang, H.~Zhang, J.~Yang, C.~Li, Y.~Zhong, L.~Wang, L.~Yuan,
  L.~Zhang, J.-N. Hwang \emph{et~al.}, ``Grounded language-image
  pre-training,'' in \emph{Proceedings of the IEEE/CVF Conference on Computer
  Vision and Pattern Recognition}, 2022, pp. 10\,965--10\,975.

\bibitem{li2022language}
B.~Li, K.~Q. Weinberger, S.~Belongie, V.~Koltun, and R.~Ranftl,
  ``Language-driven semantic segmentation,'' \emph{arXiv preprint
  arXiv:2201.03546}, 2022.

\bibitem{zhang2022pointclip}
R.~Zhang, Z.~Guo, W.~Zhang, K.~Li, X.~Miao, B.~Cui, Y.~Qiao, P.~Gao, and H.~Li,
  ``Pointclip: Point cloud understanding by clip,'' in \emph{Proceedings of the
  IEEE/CVF Conference on Computer Vision and Pattern Recognition}, 2022, pp.
  8552--8562.

\bibitem{xue2023ulip}
L.~Xue, M.~Gao, C.~Xing, R.~Mart{\'\i}n-Mart{\'\i}n, J.~Wu, C.~Xiong, R.~Xu,
  J.~C. Niebles, and S.~Savarese, ``Ulip: Learning a unified representation of
  language, images, and point clouds for 3d understanding,'' in
  \emph{Proceedings of the IEEE/CVF Conference on Computer Vision and Pattern
  Recognition}, 2023, pp. 1179--1189.

\bibitem{fan2021acr}
Z.~Fan, Z.~Song, J.~Xu, Z.~Wang, K.~Wu, H.~Liu, and J.~He, ``Acr-pose:
  Adversarial canonical representation reconstruction network for category
  level 6d object pose estimation,'' \emph{arXiv preprint arXiv:2111.10524},
  2021.

\bibitem{lin2022category}
J.~Lin, Z.~Wei, C.~Ding, and K.~Jia, ``Category-level 6d object pose and size
  estimation using self-supervised deep prior deformation networks,'' in
  \emph{European Conference on Computer Vision}.\hskip 1em plus 0.5em minus
  0.4em\relax Springer, 2022, pp. 19--34.

\bibitem{chen2021sgpa}
K.~Chen and Q.~Dou, ``Sgpa: Structure-guided prior adaptation for
  category-level 6d object pose estimation,'' in \emph{Proceedings of the
  IEEE/CVF International Conference on Computer Vision}, 2021, pp. 2773--2782.

\bibitem{zheng2023hs}
L.~Zheng, C.~Wang, Y.~Sun, E.~Dasgupta, H.~Chen, A.~Leonardis, W.~Zhang, and
  H.~J. Chang, ``Hs-pose: Hybrid scope feature extraction for category-level
  object pose estimation,'' \emph{arXiv preprint arXiv:2303.15743}, 2023.

\bibitem{umeyama1991least}
S.~Umeyama, ``Least-squares estimation of transformation parameters between two
  point patterns,'' \emph{IEEE Transactions on Pattern Analysis \& Machine
  Intelligence}, vol.~13, no.~04, pp. 376--380, 1991.

\bibitem{vaswani2017attention}
A.~Vaswani, N.~Shazeer, N.~Parmar, J.~Uszkoreit, L.~Jones, A.~N. Gomez,
  {\L}.~Kaiser, and I.~Polosukhin, ``Attention is all you need,''
  \emph{Advances in neural information processing systems}, vol.~30, 2017.

\bibitem{chen2020learning}
D.~Chen, J.~Li, Z.~Wang, and K.~Xu, ``Learning canonical shape space for
  category-level 6d object pose and size estimation,'' in \emph{Proceedings of
  the IEEE/CVF conference on computer vision and pattern recognition}, 2020,
  pp. 11\,973--11\,982.

\bibitem{xu2022groupvit}
J.~Xu, S.~De~Mello, S.~Liu, W.~Byeon, T.~Breuel, J.~Kautz, and X.~Wang,
  ``Groupvit: Semantic segmentation emerges from text supervision,'' in
  \emph{Proceedings of the IEEE/CVF Conference on Computer Vision and Pattern
  Recognition}, 2022, pp. 18\,134--18\,144.

\bibitem{wang2023res}
L.~Wang, Z.~He, R.~Dang, H.~Chen, C.~Liu, and Q.~Chen, ``Res-sts: Referring
  expression speaker via self-training with scorer for goal-oriented
  vision-language navigation,'' \emph{IEEE Transactions on Circuits and Systems
  for Video Technology}, 2023.

\bibitem{he2023learning}
Z.~He, L.~Wang, R.~Dang, S.~Li, Q.~Yan, C.~Liu, and Q.~Chen, ``Learning depth
  representation from rgb-d videos by time-aware contrastive pre-training,''
  \emph{IEEE Transactions on Circuits and Systems for Video Technology}, 2023.

\bibitem{zhu2023fine}
M.~Zhu, X.~Lin, R.~Dang, C.~Liu, and Q.~Chen, ``Fine-grained spatiotemporal
  motion alignment for contrastive video representation learning,'' \emph{arXiv
  preprint arXiv:2309.00297}, 2023.

\bibitem{zhu2022pointclip}
X.~Zhu, R.~Zhang, B.~He, Z.~Zeng, S.~Zhang, and P.~Gao, ``Pointclip v2:
  Adapting clip for powerful 3d open-world learning,'' \emph{arXiv preprint
  arXiv:2211.11682}, 2022.

\bibitem{zhang2022can}
R.~Zhang, Z.~Zeng, Z.~Guo, and Y.~Li, ``Can language understand depth?'' in
  \emph{Proceedings of the 30th ACM International Conference on Multimedia},
  2022, pp. 6868--6874.

\bibitem{xue2023ulip2}
L.~Xue, N.~Yu, S.~Zhang, J.~Li, R.~Mart{\'\i}n-Mart{\'\i}n, J.~Wu, C.~Xiong,
  R.~Xu, J.~C. Niebles, and S.~Savarese, ``Ulip-2: Towards scalable multimodal
  pre-training for 3d understanding,'' \emph{arXiv preprint arXiv:2305.08275},
  2023.

\bibitem{he2017mask}
K.~He, G.~Gkioxari, P.~Doll{\'a}r, and R.~Girshick, ``Mask r-cnn,'' in
  \emph{Proceedings of the IEEE international conference on computer vision},
  2017, pp. 2961--2969.

\bibitem{dosovitskiy2020image}
A.~Dosovitskiy, L.~Beyer, A.~Kolesnikov, D.~Weissenborn, X.~Zhai,
  T.~Unterthiner, M.~Dehghani, M.~Minderer, G.~Heigold, S.~Gelly \emph{et~al.},
  ``An image is worth 16x16 words: Transformers for image recognition at
  scale,'' \emph{arXiv preprint arXiv:2010.11929}, 2020.

\bibitem{he2020momentum}
K.~He, H.~Fan, Y.~Wu, S.~Xie, and R.~Girshick, ``Momentum contrast for
  unsupervised visual representation learning,'' in \emph{Proceedings of the
  IEEE/CVF conference on computer vision and pattern recognition}, 2020, pp.
  9729--9738.

\bibitem{dyer2014notes}
C.~Dyer, ``Notes on noise contrastive estimation and negative sampling,''
  \emph{arXiv preprint arXiv:1410.8251}, 2014.

\bibitem{jia2022visual}
M.~Jia, L.~Tang, B.-C. Chen, C.~Cardie, S.~Belongie, B.~Hariharan, and S.-N.
  Lim, ``Visual prompt tuning,'' in \emph{European Conference on Computer
  Vision}.\hskip 1em plus 0.5em minus 0.4em\relax Springer, 2022, pp. 709--727.

\bibitem{rodrigues1840lois}
O.~Rodrigues, ``Des lois g{\'e}om{\'e}triques qui r{\'e}gissent les
  d{\'e}placements d'un syst{\`e}me solide dans l'espace, et de la variation
  des coordonn{\'e}es provenant de ces d{\'e}placements consid{\'e}r{\'e}s
  ind{\'e}pendamment des causes qui peuvent les produire,'' \emph{Journal de
  math{\'e}matiques pures et appliqu{\'e}es}, vol.~5, pp. 380--440, 1840.

\bibitem{lin2021dualposenet}
J.~Lin, Z.~Wei, Z.~Li, S.~Xu, K.~Jia, and Y.~Li, ``Dualposenet: Category-level
  6d object pose and size estimation using dual pose network with refined
  learning of pose consistency,'' in \emph{Proceedings of the IEEE/CVF
  International Conference on Computer Vision}, 2021, pp. 3560--3569.

\bibitem{zhang2022ssp}
R.~Zhang, Y.~Di, F.~Manhardt, F.~Tombari, and X.~Ji, ``Ssp-pose: Symmetry-aware
  shape prior deformation for direct category-level object pose estimation,''
  in \emph{2022 IEEE/RSJ International Conference on Intelligent Robots and
  Systems (IROS)}.\hskip 1em plus 0.5em minus 0.4em\relax IEEE, 2022, pp.
  7452--7459.

\bibitem{zhou2023dr}
L.~Zhou, Z.~Liu, R.~Gan, H.~Wang, and M.~H. Ang~Jr, ``Dr-pose: A two-stage
  deformation-and-registration pipeline for category-level 6d object pose
  estimation,'' \emph{arXiv preprint arXiv:2309.01925}, 2023.

\bibitem{lin2022sar}
H.~Lin, Z.~Liu, C.~Cheang, Y.~Fu, G.~Guo, and X.~Xue, ``Sar-net: Shape
  alignment and recovery network for category-level 6d object pose and size
  estimation,'' in \emph{Proceedings of the IEEE/CVF Conference on Computer
  Vision and Pattern Recognition}, 2022, pp. 6707--6717.

\bibitem{zhang2022rbp}
R.~Zhang, Y.~Di, Z.~Lou, F.~Manhardt, F.~Tombari, and X.~Ji, ``Rbp-pose:
  Residual bounding box projection for category-level pose estimation,'' in
  \emph{European Conference on Computer Vision}.\hskip 1em plus 0.5em minus
  0.4em\relax Springer, 2022, pp. 655--672.

\bibitem{lin2020convolution}
Z.-H. Lin, S.-Y. Huang, and Y.-C.~F. Wang, ``Convolution in the cloud: Learning
  deformable kernels in 3d graph convolution networks for point cloud
  analysis,'' in \emph{Proceedings of the IEEE/CVF conference on computer
  vision and pattern recognition}, 2020, pp. 1800--1809.

\end{thebibliography}

\vspace{43pt}


\begin{IEEEbiography}[{\includegraphics[width=1in,height=1.25in,clip,keepaspectratio]{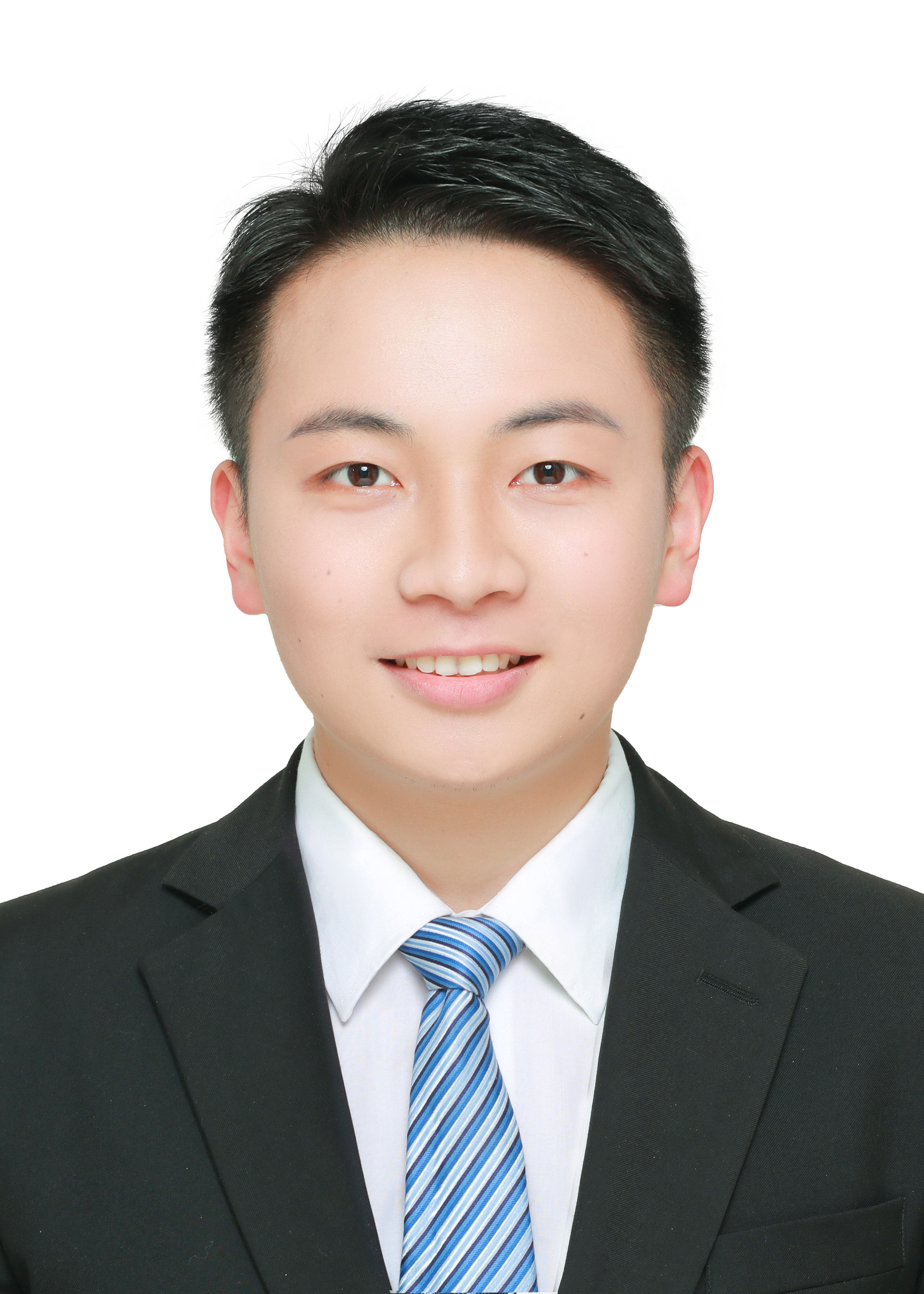}}]{Xiao Lin}
received his B.Eng. degree in Automation from Tongji University, Shanghai, China, in 2021, where he is currently pursuing his direct Ph.D. degree with the Robotics and Artificial Intelligence Laboratory. His research interests are in visual perception for robotics, with a focus on 3D vision detection and 6D object pose estimation.
\end{IEEEbiography}

\begin{IEEEbiography}[{\includegraphics[width=1in,height=1.25in,clip,keepaspectratio]{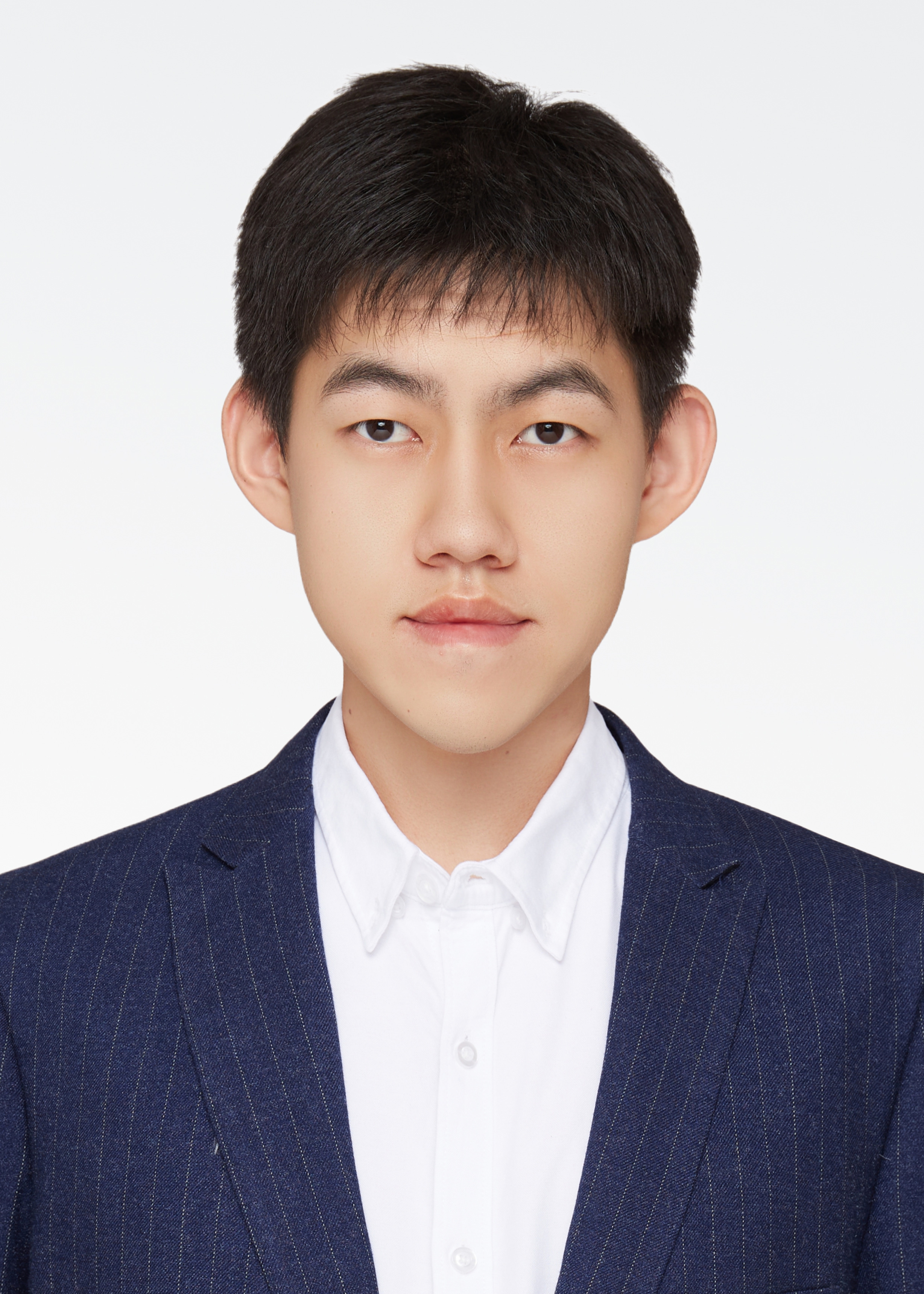}}]{Minghao Zhu}
	received the B.Eng. degree from Tongji University in 2021 and the B.Eng. degree from Polytechnic University of Turin in 2021. He is currently pursuing the Ph.D. degree with Tongji University. His research interests include computer vision, deep learning, with specific focus on large-scale and multimodal video understanding.
\end{IEEEbiography}

\begin{IEEEbiography}[{\includegraphics[width=1in,height=1.25in,clip,keepaspectratio]{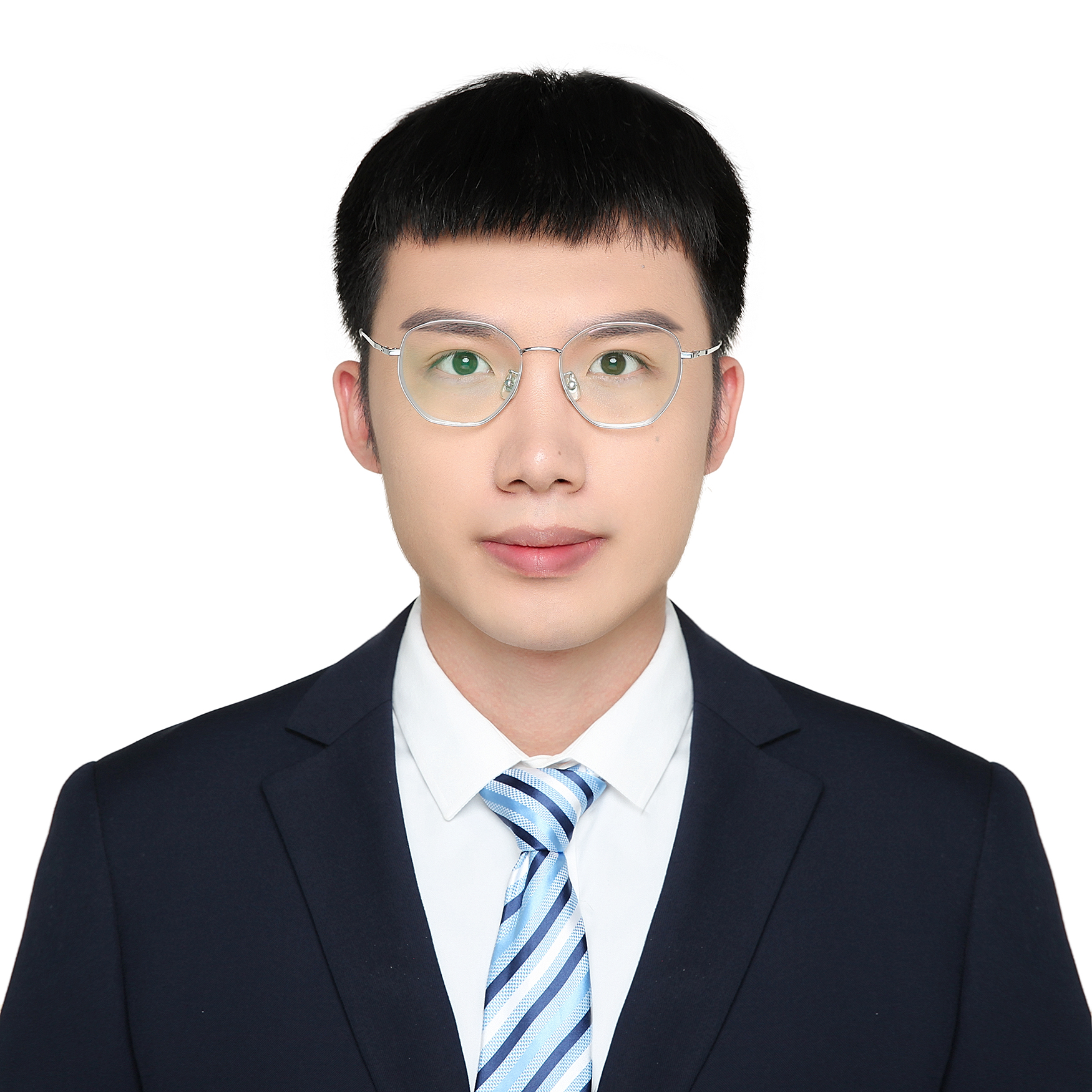}}]{Ronghao Dang}
received the B.Eng. degree from Tongji University in 2021. He is currently pursuing the M.S. degree at the Department of Control Science and Engineering, College of Electronics and Information Engineering, Tongji University. His research interests include object navigation and Multimodal large model.
\end{IEEEbiography}


\begin{IEEEbiography}[{\includegraphics[width=1in,height=1.25in,clip,keepaspectratio]{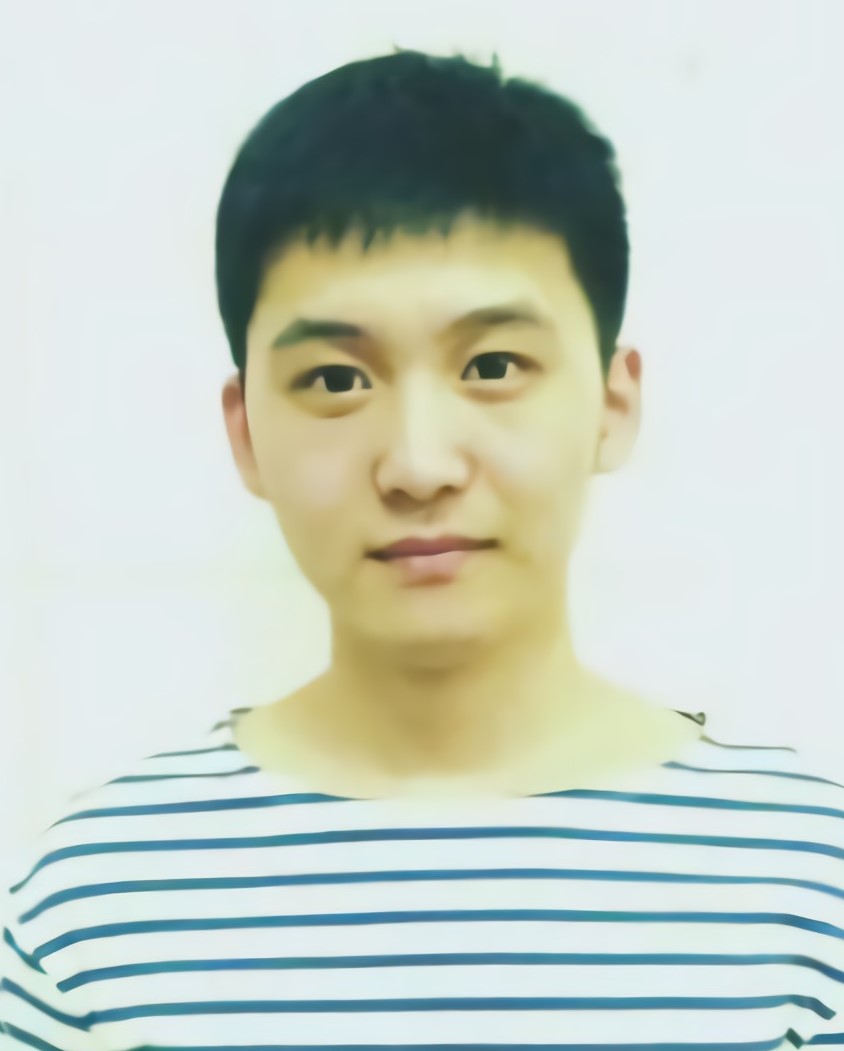}}]{Guangliang Zhou}
received his B.Eng. degree in Automation from Tongji University, Shanghai, China, in 2017, where he is currently pursuing his Ph.D. degree with the Robotics and Artificial Intelligence Laboratory. His research interests are in visual perception for robotics, with a focus on 6D object pose estimation and grasp detection.
\end{IEEEbiography}

\begin{IEEEbiography}[{\includegraphics[width=1in,height=1.25in,clip,keepaspectratio]{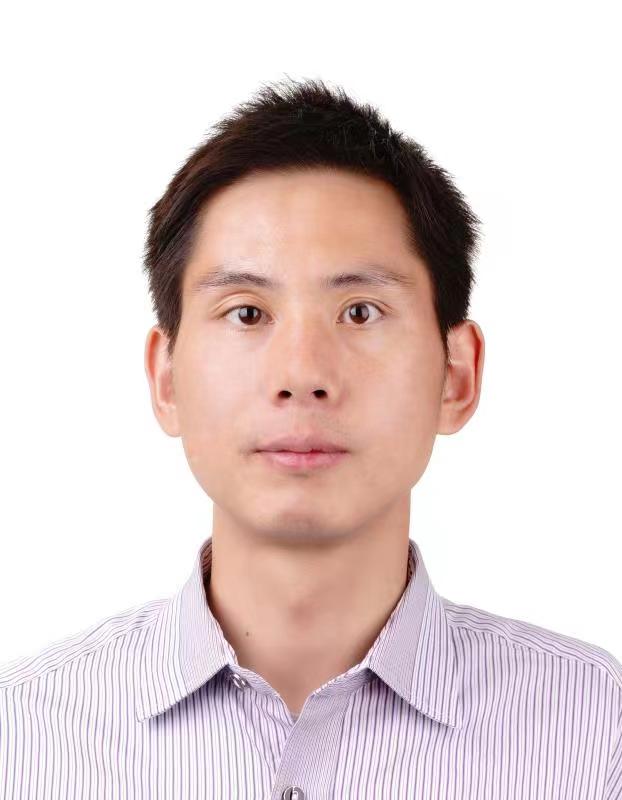}}]{Shaolong Shu} (Senior Member, IEEE)
received his B.Eng. degree in automatic control, and his Ph.D. degree in control theory and control engineering from Tongji University, Shanghai, China, in 2003 and 2008, respectively. Since July, 2008, he has been with the School of Electronics and Information Engineering, Tongji University, Shanghai, China, where he is currently a full professor. From August, 2007 to February, 2008 and from April, 2014 to April, 2015, he was a visiting scholar in Wayne State University, Detroit, MI, USA. His main research interests include state estimation and control of discrete event systems and cyber–physical systems.
\end{IEEEbiography}

\begin{IEEEbiography}[{\includegraphics[width=1in,height=1.25in,clip,keepaspectratio]{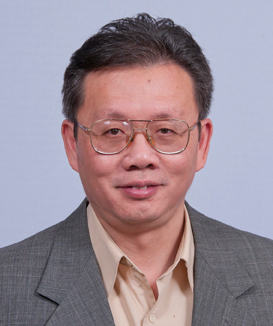}}]{Feng Lin} (Fellow, IEEE)
received his B.Eng. degree in electrical engineering from Shanghai Jiao-Tong University, Shanghai, China, in 1982, and his M.A.Sc. and Ph.D. degrees in electrical engineering from the University of Toronto, Toronto, Canada, in 1984 and 1988, respectively. From 1987 to 1988, he was a postdoctoral fellow at Harvard University, Cambridge, MA. Since 1988, he has been with the Department of Electrical and Computer Engineering, Wayne State University, Detroit, Michigan, where he is currently a professor. His research interests include discrete-event systems, hybrid systems, robust control, and their applications in alternative energy, biomedical systems, and automotive control. He is the author of a book entitled Robust Control Design: An Optimal Control Approach. He was a consultant for GM, Ford, Hitachi and other auto companies. He co-authored a paper that received a George Axelby outstanding paper award from IEEE Control Systems Society. He was an associate editor of IEEE Transactions on Automatic Control.
\end{IEEEbiography}

\begin{IEEEbiography}[{\includegraphics[width=1in,height=1.25in,clip,keepaspectratio]{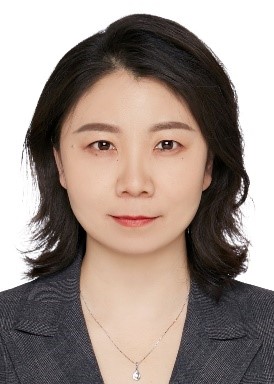}}]{Chengju Liu}
received the Ph.D. degree in control theory and control engineering from Tongji University, Shanghai, China, in 2011. From October 2011 to July 2012, she was with the BEACON Center, Michigan State University, East Lansing, MI, USA, as a Research Associate. From March 2011 to June 2013, she was a Postdoctoral Researcher with Tongji University, where she is currently a Professor with the Department of Control Science and Engineering, College of Electronics and Information Engineering, and a Chair Professor of Tongji Artificial Intelligence (Suzhou) Research Institute. She is also a Team Leader with the TJArk Robot Team, Tongji University. Her research interests include intelligent control, motion control of legged robots, and evolutionary computation.
\end{IEEEbiography}

\begin{IEEEbiography}[{\includegraphics[width=1in,height=1.25in,clip,keepaspectratio]{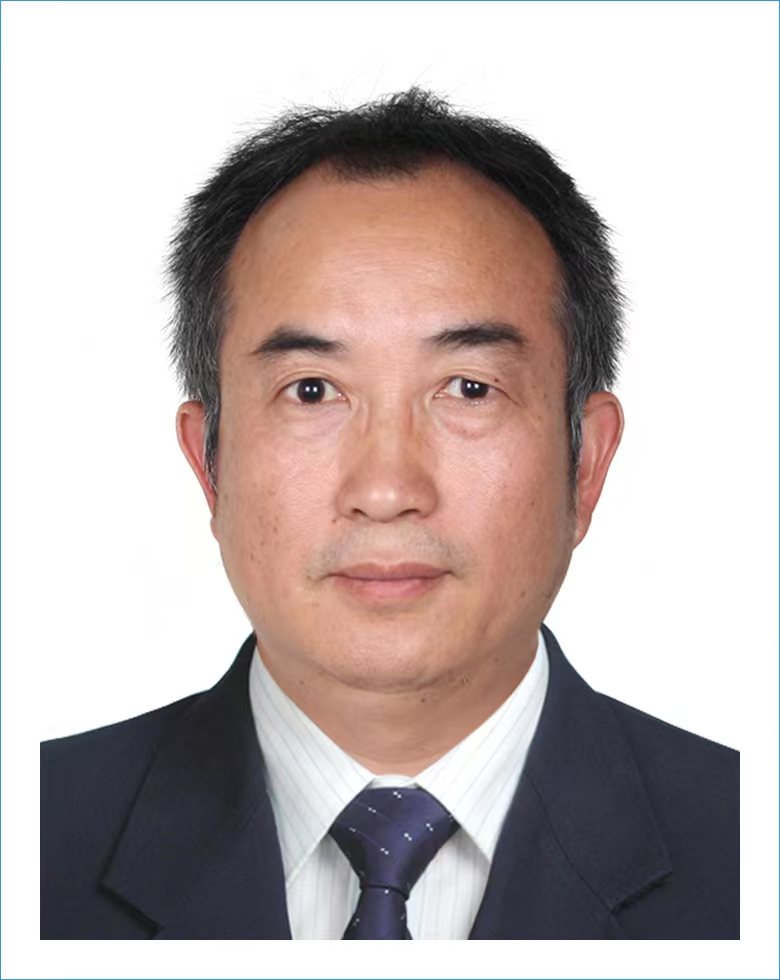}}]{Qijun Chen} (Senior Member, IEEE)
received the B.S. degree in automation from Huazhong University of Science and Technology, Wuhan, China, in 1987, the M.S. degree in information and control engineering from Xi’an Jiaotong University, Xi’an, China, in 1990, and the Ph.D. degree in control theory and control engineering from Tongji University, Shanghai, China, in 1999. He is currently a Full Professor in the College of Electronics and Information Engineering, Tongji University. His research interests include robotics control, environmental perception, and understanding of mobile robots and bioinspired control.
\end{IEEEbiography}

\vfill

\end{document}